\documentclass[pdflatex,sn-mathphys-num]{sn-jnl} 
\usepackage{graphicx}%
\usepackage{multirow}%
\usepackage{amsmath,amssymb,amsfonts}%
\usepackage{amsthm}%
\usepackage{mathrsfs}%
\usepackage[title]{appendix}%
\usepackage{xcolor}%
\usepackage{textcomp}%
\usepackage{manyfoot}%
\usepackage{booktabs}%
\usepackage{adjustbox}
\usepackage{algorithm}%
\usepackage{algorithmicx}%
\usepackage{algpseudocode}%
\usepackage{subcaption}
\usepackage{array}
\usepackage{listings}%

\usepackage[vietnamese]{babel}
\addto\captionsvietnamese{\renewcommand{\tablename}{Table}}
\addto\captionsvietnamese{\renewcommand{\figurename}{Figure}}
\addto\captionsvietnamese{\renewcommand{\abstractname}{Abstract}}

\usepackage{booktabs}
\usepackage{array}
\usepackage{xcolor}
\usepackage{colortbl}
\usepackage{siunitx}
\definecolor{best}{HTML}{2E8B57}
\definecolor{second}{HTML}{4682B4}
\definecolor{third}{HTML}{DAA520}
\definecolor{headercolor}{HTML}{4A90E2}
\definecolor{groupcolor}{HTML}{F0F0F0}

\usepackage[utf8]{inputenc}
\theoremstyle{thmstyleone}%
\theoremstyle{thmstyletwo}%

\theoremstyle{thmstylethree}%

\begin{document}

\title[Towards Signboard-Oriented Visual Question Answering: ViSignVQA Dataset and Benchmark]{Towards Signboard-Oriented Visual Question Answering: ViSignVQA Dataset, Method and Benchmark}



\author[1,2]{\fnm{Hieu} \sur{Minh Nguyen}}\email{22520440@gm.uit.edu.vn}

\author[1,2]{\fnm{Tam} \sur{Le-Thanh Dang}}\email{22521290@gm.uit.edu.vn}

\author*[1,2]{\fnm{Kiet} \sur{Van Nguyen}}\email{kietnv@uit.edu.vn}

\affil[1]{\orgname{University of Information Technology, Ho Chi Minh City, Vietnam}}

\affil[2]{\state{Vietnam National University}, \country{Ho Chi Minh City, Vietnam}}

\abstract{Understanding signboard text in natural scenes is essential for real-world applications of Visual Question Answering (VQA), yet remains underexplored, particularly in low-resource languages. We introduce ViSignVQA, the first large-scale Vietnamese dataset designed for signboard-oriented VQA, which comprises 10,762 images and 25,573 question-answer pairs. The dataset captures the diverse linguistic, cultural, and visual characteristics of Vietnamese signboards, including bilingual text, informal phrasing, and visual elements such as color and layout. To benchmark this task, we adapted state-of-the-art VQA models (e.g., BLIP-2, LaTr, PreSTU, and SaL) by integrating a Vietnamese OCR model (SwinTextSpotter) and a Vietnamese pretrained language model (ViT5). The experimental results highlight the significant role of the OCR-enhanced context, with F1-score improvements of up to 209\% when the OCR text is appended to questions. Additionally, we propose a multi-agent VQA framework combining perception and reasoning agents with GPT-4, achieving 75.98\% accuracy via majority voting. Our study presents the first large-scale multimodal dataset for Vietnamese signboard understanding. This underscores the importance of domain-specific resources in enhancing text-based VQA for low-resource languages. ViSignVQA serves as a benchmark capturing real-world scene text characteristics and supporting the development and evaluation of OCR-integrated VQA models in Vietnamese.}

\maketitle
\section{Introduction}\label{sec1}
Over the past few years, swift progress in technology and science has led to significant progress across various fields, particularly in Artificial Intelligence (AI). Researchers in both Computer Vision (CV) and Natural Language Processing (NLP) have focused on developing multimodal models that combine the ability to understand and respond to tasks involving both visual and linguistic aspects. Examples of such multimodal models include OpenAI’s CLIP\cite{clip}, which aligns images and text for improved image recognition, and Google’s VLMo\cite{vlmo}, a unified model that performs tasks across vision, language, and multimodal data. These models exemplify the integration of multiple modalities to enhance AI’s capabilities in processing and generating information on different types of input. This, in turn, has catalyzed the advancement of research in Visual Question Answering (VQA), a domain in which models are designed to automatically interpret and answer questions based on image content.

In the last few years, research on VQA in English has increased remarkably, largely because of the abundance of English-language resources. Additionally, most large language models (LLMs) tend to perform exceptionally well in English, further contributing to this increase. We have witnessed the emergence of numerous English-language VQA datasets, such as VQA V1\cite{vqav1}, VQA V2\cite{vqav2}, and TextVQA\cite{textvqa}. However, in the specific domain of integrating Optical Character Recognition (OCR) with Visual Question Answering (VQA), the available datasets remain limited. Currently, OCR-VQA-200k\cite{ocrvqa} is the only dataset that focuses on OCR-VQA in English; however, its images are primarily book covers, with questions centered around the information on these covers. To date, there has been little research conducted on this dataset. Moreover, there is currently no publicly available dataset for answering questions related to text on signboards.

When applied to signboard VQA, this task requires an even more nuanced and adaptable approach, necessitating a sophisticated combination of image processing and natural language understanding. The model must accurately recognize and interpret text from various signboards, which often involves dealing with diverse fonts, languages, and layouts. In addition to identifying visual content, the model must comprehend the context and intent behind the question posed. This includes understanding references to specific text elements, interpreting directional information, and discerning meanings based on the surrounding visual cues.

The challenge is heightened by the need for the model to generate answers that are both contextually accurate and relevant to the specific query. This requires the model to seamlessly integrate OCR capabilities with VQA and synthesize information from different modalities. The ability to effectively handle these complexities necessitates a deep integration of natural language processing, computer vision, and OCR, all working in harmony to interpret and respond to the nuanced demands of signboards VQA tasks.

In the realm of Vietnamese VQA, Tran et al. \cite{tran2021vivqa} introduced ViVQA, the initial dataset crafted for the Vietnamese Visual Question Answering task. ViVQA is a smaller-scale version of the VQA v2 dataset, created using semi-automatic methods. Nevertheless, this dataset was deemed insufficiently comprehensive. To overcome the shortcomings of the ViVQA dataset, Nguyen et al. \cite{openvivqa} introduced OpenViVQA, the first comprehensive, hand-annotated dataset for VQA in Vietnamese. OpenViVQA comprises more than 11,000 images paired with over 37,000 question-answer pairs. Unlike earlier datasets, OpenViVQA’s shift to open-ended, natural language answers, from single words to sentences, departs from categorical responses, offering both challenges and opportunities for VQA development.

A notable limitation of the ViVQA dataset is that it does not address scene text. In contrast, the OpenViVQA dataset incorporates this element as a major challenge, with approximately half of its samples involving text in the visual scene, thereby reflecting real-world vision–language scenarios more effectively. Despite its advancements, OpenViVQA is not without its limitations, which led to the creation of the ViTextVQA \cite{nguyen2024vitextvqa} dataset. The ViTextVQA dataset comprises 16,762 images and 50,342 question-answer pairs, with a primary emphasis on extracting information from both text and scene text within images. This dataset was specifically designed to address the challenges that traditional VQA models often face when dealing with textual content. ViTextVQA stands out not only for its diversity but also as a crucial resource for assessing and enhancing VQA models that are adept at processing text, particularly in Vietnamese.
\begin{figure}[htp]
    \centering
    \includegraphics[width=1\textwidth]{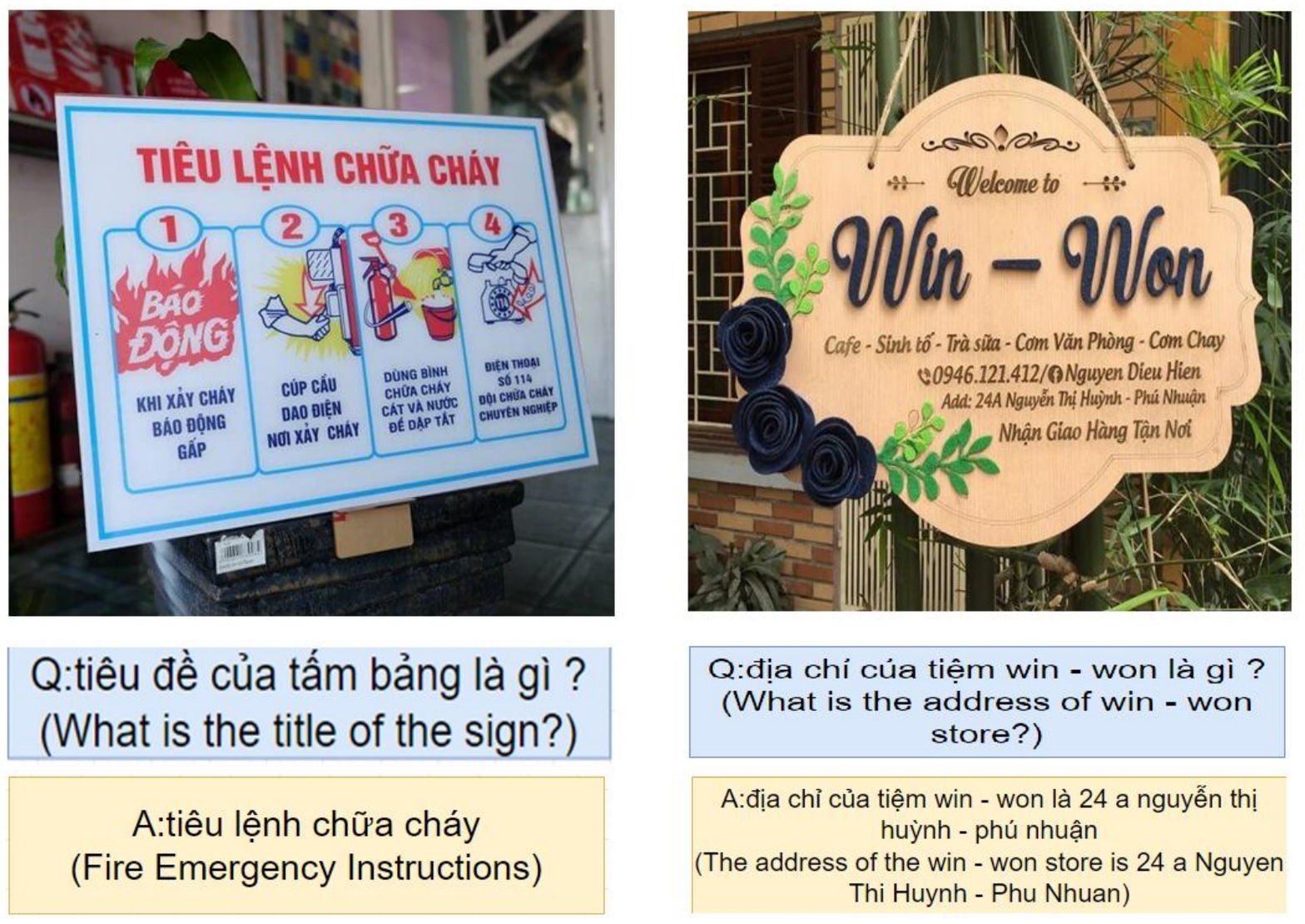}
    \caption{Examples Retrieved from the ViSignVQA Dataset.}
    \label{vidu}
\end{figure}

However, similar to English, the availability of datasets that integrate both OCR and VQA in Vietnamese is limited. Currently, the ViOCRVQA\cite{pham2025viocrvqa} dataset, which is the Vietnamese version of OCR-VQA-200k, primarily focuses on OCR-based question answering related to book covers and lacks context diversity. To address this gap, we developed a novel dataset called Vietnamese Signboard Visual Question Answering (ViSignVQA). This dataset comprises 10,762 images and 25,573 question-answer pairs, specifically designed to address questions related to the content of signboards. ViSignVQA includes a diverse range of signboard designs, lettering, and surrounding contexts, thereby offering a significant enhancement to the available OCR-VQA data resources in Vietnamese.

Our main contributions are as follows:
\begin{itemize}
    \item We created the first high-quality, large-scale ViSignVQA dataset tailored for Visual Question Answering on Vietnamese signboards, emphasizing content and context specific to signboards.
    \item We proposed a novel multi-agent Visual Question Answering (VQA) architecture integrated with Optical Character Recognition (OCR) and evaluated its effectiveness using large language models (LLMs) combined with perception modules for Vietnamese signboard understanding on the ViSignVQA dataset.
    \item We analyzed the challenges of the ViSignVQA dataset by assessing the performance of OCR systems in extracting text from signboard images.
    \item We conducted experiments on the ViSignVQA dataset using several state-of-the-art models for text detection and recognition in real-world signboard scenarios.
    \item We performed a survey and comparison of prompt-based fine-tuning methods to enhance the performance of large language models for Vietnamese language tasks.
\end{itemize}
.

\section{Related work}\label{sec2}
\subsection{From traditional to OCR in visual question answering}
\subsubsection{Former Visual Question Answering Datasets}
The Visual Question Answering (VQA) task continues to pose significant challenges to multimodal models and offers substantial research and practical value. A major milestone was the release of the VQA v1 dataset by Antol et al., which was largely constructed from MS COCO \cite{mscoco}. VQA v1 spurred the development of many new evaluation strategies and modeling approaches. Teney et al. \cite{teney2018visual} later critiqued VQA v1, noting that many answers could be addressed via classification instead of open-ended generation. To overcome this drawback, Goyal et al. \cite{goyal2017making} introduced VQA v2, pairing consistent questions with diverse answers across images to balance the frequency of responses for various question types.
Motivated by the success of Visual Question Answering (VQA) research in English, numerous studies have developed and applied VQA datasets, models, and evaluation protocols for the Vietnamese language.
\begin{table}[htp]
\centering
\caption{Comparative overview of established OCR-VQA datasets.}
\begin{tabular}{lcrrr}
\hline
\textbf{Dataset}            & \textbf{Language}    & \textbf{Images}             & \textbf{Questions}            & \textbf{Answers}              \\ \hline
TextVQA \cite{textvqa}                    & English                   & 28,408                      & 45,336                        & 453,360                       \\
DocVQA \cite{mathew2021docvqa}                     &                      & 12,767                      & 50,000                        & 50,000                       \\
OCR-VQA-200k \cite{ocrvqa}                    &                      & 207,572                     & 1,002,146                     & 1,002,146                     \\
InfographicVQA \cite{mathew2022infographicvqa}             &                      & 5,485                      & 30,035                        & 30,035                        \\
VisualMRC \cite{zhang2022visualmrc}                  &                      & 10,197                      & 30,562                        & 30,562                        \\
VizWiz \cite{gurari2018vizwiz}                     &                      & 32,842                      & 265,420                       & 265,420                       \\ \hline
EVJVQA \cite{evjvqa} & English, Vietnamese, Japanese & 4909 & 33,790 & 33,790 \\ \hline

ViVQA \cite{tran2021vivqa}                      & Vietnamese           & 10,328                      & 15,000                        & 15,000                        \\
OpenViVQA \cite{openvivqa}                  &                      & 11,199                      & 21,271                        & 21,271                        \\
ViOCRVQA \cite{pham2025viocrvqa}                    &                      & 28,282                      & 123,781                       & 123,781                       \\
ViTextVQA \cite{nguyen2024vitextvqa} &                      & 16,762                     & 50,342                        & 50,342                        \\ \hline

\textbf{ViSignVQA (Ours)} & Vietnamese & \textbf{10,762} & \textbf{25,573} & \textbf{25,573} \\ \hline
\end{tabular}%
\label{table2}
\end{table}

\textbf{ViVQA dataset}: The ViVQA dataset marks a significant milestone in the development of VQA tasks in Vietnamese, being the first publicly released VQA dataset in this language. Comprising 10,320 images and 15,000 question–answer pairs, ViVQA provides a sufficiently large resource for research and development in Vietnamese VQA. However, the dataset is relatively simple, with the answers primarily taking the form of classifications.

\textbf{EVJVQA dataset}: Nguyen et al. made a significant contribution to the field of VQA by introducing the EVJVQA dataset, which not only explores the challenges of multilingual VQA but also takes into account the cultural nuances specific to Vietnam. The dataset's design allows researchers to investigate how VQA systems perform across different languages and cultural contexts, which is crucial for developing more inclusive and globally applicable AI models. With over 33,000 question-answer pairs in Vietnamese, English, and Japanese, linked to approximately 5,000 images taken in Vietnam, the EVJVQA dataset provides a rich resource for studying the intersection of language, culture, and visual understanding.

The importance of the EVJVQA dataset is reinforced by its designation as the official benchmark for the general VQA task in the Multilingual Visual Question Answering shared task at the 9th Workshop on Vietnamese Language and Speech Processing (VLSP 2022\footnote{\href{https://vlsp.org.vn/vlsp2022/eval/evjvqa}{https://vlsp.org.vn/vlsp2022/eval/evjvqa}}
). Serving as a standardized evaluation resource, the dataset plays a critical role in advancing research and development, enabling rigorous comparison of VQA models within a multilingual and culturally contextualized framework.

\textbf{OpenViVQA dataset: }The ViVQA dataset, developed by Tran et al. \cite{tran2021vivqa}, mainly approaches Vietnamese VQA as a task of selecting or classifying answers. Recognizing the limitations of this approach, Nguyen et al. \cite{openvivqa} introduced the OpenViVQA dataset, which is the first extensive VQA dataset in Vietnamese that features open-ended answers. OpenViVQA consists of 11,199 images and 37,914 question–answer pairs, in which the answers are provided in multiple natural language forms (words, phrases, or full sentences), offering greater expressiveness by avoiding the limitations of predefined answer classes. Consequently, this dataset is unsuitable for classification-based experiments, given its open-ended nature.
\subsubsection{OCR with Visual Question Answering Datasets}
Traditional VQA approaches have faced notable limitations in addressing questions that require reading and reasoning over textual content embedded within images. To overcome this challenge, the TextVQA dataset was introduced, in which the questions explicitly depend on the text appearing in the images. To address this challenge, the LoRRA model \cite{textvqa} was proposed, incorporating Optical Character Recognition (OCR) to extract and leverage textual information as part of the answer generation process. In parallel, the OCR-VQA-200k dataset was released, comprising more than 200,000 book cover visuals and over 1 million associated sets of questions with their corresponding answers, thereby substantially advancing the development of OCR-based VQA models.

In the Vietnamese context, localized versions of these datasets ViTextVQA and ViOCR-VQA have been developed. ViTextVQA includes more than 16,000 images and over 50,000 QA pairs. Extensive experimentation with state-of-the-art models revealed that the sequence in which OCR tokens are processed and selected plays a critical role in formulating accurate answers. Meanwhile, ViOCR-VQA contains over 28,000 images and 120,000 question–answer pairs, focusing specifically on answering questions about textual content found on Vietnamese book covers.

However, OCR applications in everyday life particularly in Vietnamese commonly involve signboards, yet no dataset has been developed to address question answering based on signboard images. This gap motivated us to create and release the ViSignVQA dataset.

\subsection{OCR-Integrated Approaches for Visual Question Answering}
\label{related_methods}
Despite the emergence of numerous advanced approaches, the Visual Question Answering (VQA) task remains a substantial challenge for both the computer vision (CV) and natural language processing (NLP) communities. Given an image and a natural language question as input, a VQA model must generate an answer by effectively integrating visual features with linguistic representations. This challenge becomes even more complex when incorporating Optical Character Recognition (OCR) into VQA models, enabling them to read text within images and reason about it to generate accurate answers.

Currently, in the realm of integrating OCR into VQA, one of the most prominent models is LoRRA. This model was developed following the release of the TextVQA dataset by Amanpreet Singh and colleagues. The LoRRA model is specifically designed to read text embedded within images, reason about this information in relation to both the visual content and the accompanying question, and generate an answer. This answer may involve either direct utilization of textual strings detected in the image or deductions that integrate both textual and visual cues. Demonstrated to surpass previous state-of-the-art VQA models on the TextVQA dataset, LoRRA underscores the effectiveness of incorporating OCR for addressing the challenges inherent in OCR-based VQA tasks.

In Vietnam, Pham et al. introduced a novel approach called Vision-Reader, which integrates several OCR systems, including Swintext Spotter, and a pretrained language model in conjunction with the release of the ViOCRVQA dataset to address its challenges. This approach achieved an Exact Match (EM) score of 0.4116 and an F1-score of 0.6990 on the test set, demonstrating a new direction in tackling VQA tasks by integrating OCR within the Vietnamese context. This development highlights the potential for further advancements in OCR-VQA models tailored for the Vietnamese language.

\subsection{LLM based Methods}
Recent progress in Large Language Models (LLMs), exemplified by GPT-4\cite{openai2023gpt4}, PaLM\cite{chowdhery2022palm}, and LLaMA\cite{touvron2023llama}, have significantly influenced multimodal reasoning tasks, including Visual Question Answering (VQA). These models, originally designed for text-based tasks, have been increasingly integrated with visual encoders to process and reason over images. Approaches like BLIP-2\cite{li2023blip}, MiniGPT-4\cite{zhu2023minigpt4}, and LLaVA\cite{liu2023llava} exemplify this integration by employing pre-trained vision encoders and aligning their outputs with LLMs through lightweight adapters or multimodal projection layers. These models demonstrate strong zero-shot and few-shot performance on various VQA benchmarks without requiring extensive task-specific fine-tuning.

In the context of signboard understanding, where textual content embedded in images plays a crucial role, LLM-based methods are particularly promising. Methods such as Kosmos-2\cite{peng2023kosmos2} have shown the potential of integrating OCR signals and spatial context into large-scale vision-language pretraining, thereby enhancing the model's ability to extract and reason over scene text. However, most existing LLM-based VQA models are trained and evaluated on general-purpose datasets like VQAv2, GQA\cite{hudson2019gqa}, and OK-VQA\cite{marino2019okvqa}, which lack domain-specific challenges present in signboard images, such as multi-language text, complex backgrounds, and layout-dependent semantics.

\subsection{Multiagent Methods}
In 2025, with the rapid rise of AI agents—autonomous artificial intelligence systems capable of perceiving, reasoning, and making decisions based on human-defined goals—there has been growing interest in leveraging these agents to overcome the inherent limitations of large language models (LLMs) in complex multi-modal tasks such as Visual Question Answering (VQA). Traditional VQA systems often rely on end-to-end architectures that process both image and textual input using a unified model. However, LLMs, even when integrated with vision encoders, still face several key challenges:
\begin{itemize}
    \item Difficulty in precise object detection and fine-grained counting.
    \item Limited capacity to ground textual reasoning in visual context.
    \item A tendency to produce plausible but hallucinated answers in open-world scenarios.
\end{itemize}

To address these issues, Bowen Jiang et al\cite{jiang2024multi} recently proposed a multi-agent architecture for VQA, marking a significant shift from monolithic models to modular collaborative systems. Their method explores the zero-shot performance of foundation models in Visual Question Answering tasks, without requiring fine-tuning on domain-specific datasets. Instead of depending solely on a single model’s capabilities, their system orchestrates a set of specialized agents, each equipped to handle distinct subtasks—such as object detection, text recognition, and reasoning.

This agent-based design allows for explicit modularization, improved interpretability, and flexibility in integrating external tools (e.g., object detectors, or external knowledge sources). Preliminary experiments under zero-shot settings show promising results, particularly in tasks that involve counting and object grounding, which are traditionally weak points for standard LLMs. The authors also identify common failure modes and outline future directions, suggesting that AI agent collaboration could be a promising path forward in overcoming current VQA limitations and building more robust, generalizable visual-language systems.


\section{ViSignVQA Dataset}\label{sec3}

A comprehensive process of data acquisition and annotation was undertaken over a span of six weeks, with contributions from two native Vietnamese speakers. The overall methodology employed in constructing the dataset is elaborated in Section \ref{sec3.1}. Furthermore, Section \ref{sec3.2} provides a detailed examination of the dataset’s salient properties, distilled from the collection and labeling phases, with particular attention to the dimensions of questions, answers, and objects.

\subsection{Dataset Creation}
\label{sec3.1}
\begin{figure}[h]
    \centering
    \includegraphics[width=0.8\textwidth]{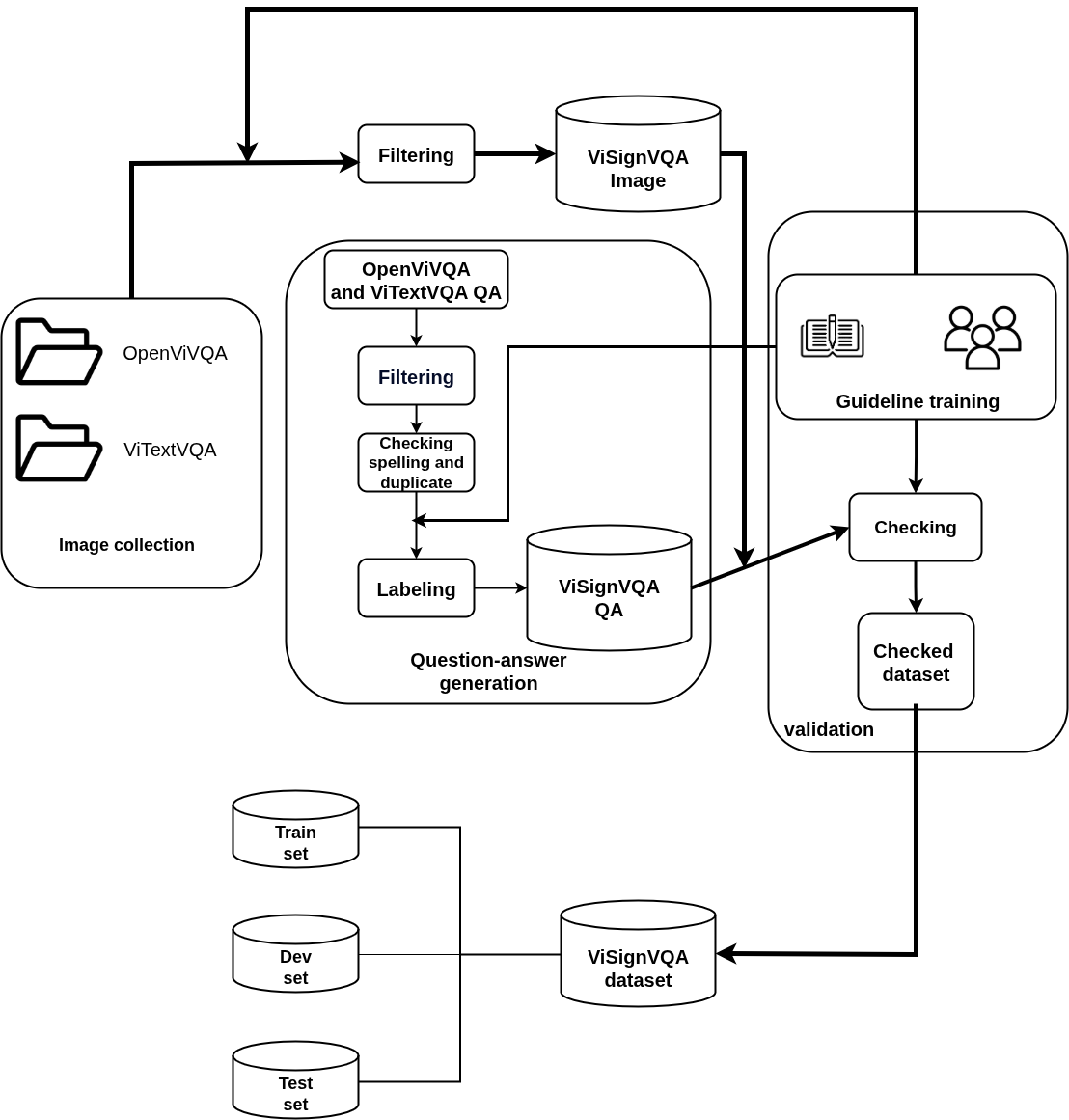}
    \caption{ViSignVQA Dataset Creation Overview.}
    \label{data_creation}
\end{figure}

To identify the most effective strategy for creating a Vietnamese signboard dataset, we evaluated multiple approaches, as shown in Figure \ref{data_creation}. We began by thoroughly assessing the scope of the task to ensure a deep understanding. Following this, we sought out datasets that would best meet our requirements and selected the OpenViVQA and ViTextVQA datasets. These datasets were chosen because they include a wealth of images with signboards and related questions. Therefore, we are confident that these two datasets are well-suited and sufficiently robust for our purposes.

Our dataset construction followed a three-phase pipeline: (1) collecting images, (2) question–answer generation, and (3) dataset validation. This procedure was inspired by the methodology employed in the creation of the ViVQA dataset, while being tailored to the characteristics of our task. A detailed description of each phase is provided below.

\subsubsection{Collecting images}
First, we collected a list of keywords associated with signboard images, including terms such as "Biển" and "ghi". We then utilized an OCR model named SwinTextSpotter, which demonstrates exceptional performance in Vietnamese text recognition, to extract all the words present in the images. Following this, we filtered the images using the following criteria:
\begin{itemize}
  \item \textbf{Rule 1:} Images containing questions with any of the predefined keywords.
  \item \textbf{Rule 2:} Images where the answer includes a word extracted by the OCR model from the image.
\end{itemize}

The rationale behind these rules is based on our observation that images containing signboards typically have questions focused on the content displayed on the signboards, which often include keywords such as "Biển" and "ghi." Additionally, since the questions are concentrated on the signboard's content, it is common for the OCR-extracted words from the image to appear in the corresponding answers.

Subsequently, we manually filtered out images that did not contain signboards, Signboards not in Vietnamese language, had signboards that were too small, or where the content on the signboards was not visible, with the following exceptions:
\begin{itemize}
  \item Banners or ribbons.
  \item Signboards or content displayed on vehicles or other moving objects.
\end{itemize}

By the end of the process, we collected more than 14,000 images containing signboards from various locations in Vietnam. However, despite our efforts, the data collection process inevitably included noise and unwanted images. For duplicate images, annotators were encouraged to delete those they had already filtered.

Ultimately, we obtained over 10,762 high-quality images, a sufficient quantity for a robust dataset.

\subsubsection{Question-Answer Creation}
In this study, we have leveraged pre-existing datasets, specifically OpenViVQA and ViTextVQA, to streamline the process of question-answer creation. Given that these datasets have successfully generated a substantial volume of questions related to signboards, it was unnecessary to create entirely new questions from scratch. Instead, we focused on refining and filtering the existing questions to ensure their relevance and quality within our specific context.

In this dataset, we introduced an additional label called "is\_related," which serves to verify whether the question and answer are related to the signboard. This label takes a value of 1 if the QA pair is related to the signboard and 0 if it is not. Due to limited resources and to facilitate the testing of certain methods in future experiments, we applied this labeling only to the development set.

\subsection*{Rules for assigning value to the label "is\_related"}
The annotation guidelines for assigning the value of “is\_related” were designed to ensure correctness and consistency, thereby facilitating more reliable dataset construction and simplifying subsequent model processing. The rules are defined as follows:
\begin{itemize}
  \item \textbf{Rule 1:} If the question and answer directly pertain to the content on the signboard, the label is assigned a value of 1.
  \item \textbf{Rule 2:} If the answer is inferred from the surrounding context of one or more signboards, the label is assigned a value of 1.
  \item \textbf{Rule 3:} If the question and answer fall outside of these two criteria, the label is assigned a value of 0.
\end{itemize}
Next, Tam, a co-author of the paper, was trained to gain an understanding of the annotation guidelines and was subsequently assigned to annotate about 100 QA pairs from the development set. After confirming that Tam had correctly understood and applied the guidelines, Hieu was designated as the annotation supervisor. The supervisor's role is to monitor and support the annotator, ensuring that Tam follows the required processes and maintains the necessary quality standards.

During the annotation phase, we also conducted a validation procedure, as outlined in Section \ref{validation}. In cases where the annotator, Tam, did not fully adhere to the established requirements, additional training was provided to reinforce compliance with the guidelines. To minimize errors and improve consistency, illustrative examples of incorrectly annotated samples were supplied as corrective references.

Finally, we completed more than 2,400 samples. This significant effort has provided a solid foundation for the development and evaluation of our models, ensuring that the dataset is both comprehensive and representative of the challenges involved in signboard-related QA tasks.

\subsubsection{Validation}
\label{validation}
To measure the validation, supervisor Hieu Minh Nguyen, who has the best understanding of the guidelines, meticulously annotated the entire development set for related signboard QA pairs. We then compared Hieu's annotations with those of another annotator. We chose to use the F1-score as our evaluation metric because the number of QA pairs not related to signboards is minimal, and we aim to penalize incorrect labels more heavily. Thus, the F1-score is an appropriate metric in this context.

In this evaluation, the F1-score is calculated by comparing the related labels predicted by the annotators with the ground-truth labels provided by the annotation supervisor. Precision is defined as the proportion of correctly predicted related labels to the total number of related labels predicted by the annotators, while recall is defined as the proportion of correctly predicted related labels to the total number of related labels in the ground truth. The formula for computing the F1-score is presented in Section \ref{sec:f1}. To ensure annotation quality, we required annotators to achieve an F1-score exceeding 70\%.




By using the F1-score as a measurement tool for Inter-Annotator Agreement, we achieved a score of 98.94\%. This high score indicates a strong level of consistency between the annotators and the annotation supervisor, reflecting a thorough understanding of the guidelines and accurate application of the annotation process. Such a result suggests that the annotations are highly reliable, which is crucial for ensuring the validity of subsequent analyses and model training processes. The near-perfect agreement underscores the effectiveness of the guidelines provided and the annotators' adherence to them.

After the validation stage, the dataset was partitioned into three subsets: training, development, and test, following a 7:1:2 ratio. This partitioning facilitates more convenient analysis and experimentation.

\subsection{Dataset Exploration}
\label{sec3.2}
\begin{figure}[!ht]
    \centering
    \includegraphics[width=1\textwidth]{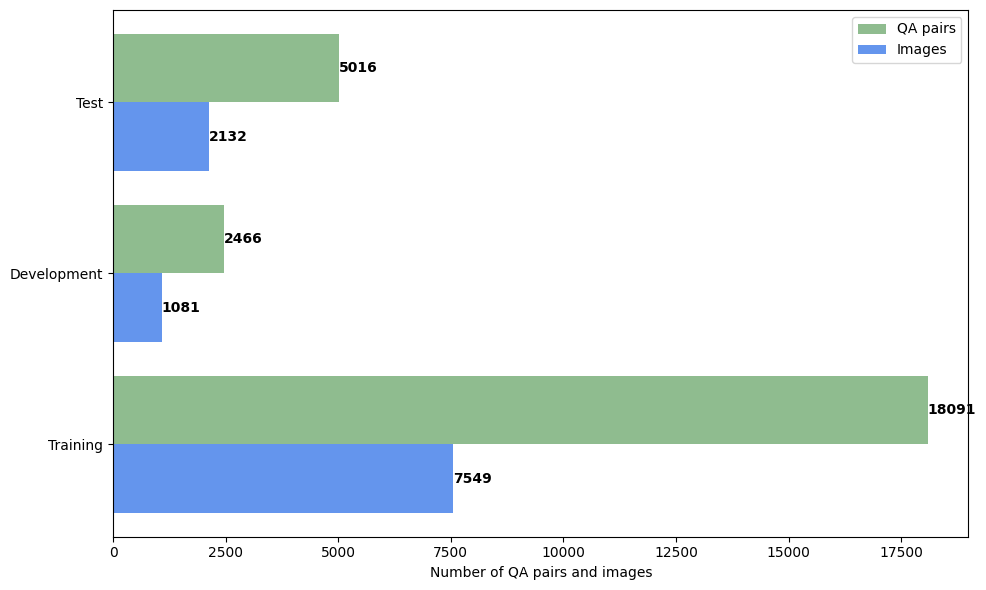}
    \caption{Distribution of Images and Question–Answer Pairs.}
    \label{num_ann_img}
\end{figure}

The ViSignVQA dataset comprises 10,762 images paired with 25,573 question-answer (QA) pairs. Figure \ref{num_ann_img} illustrates the distribution of the training, development, and test subsets within the dataset. The collection encompasses a diverse range of visual signboards, accompanied by scene text and their corresponding question–answer pairs. Each image is paired with several QA pairs, and in the development set, each QA pair is labeled with a tag indicating its relevance to the signboard. This labeling provides valuable insights into scene text and signboard detection, which are essential for accurate model predictions, as the answers may or may not pertain to the context of the signboard, adding complexity to the VQA task. Additionally, we performed statistical comparisons with other popular datasets in this field, with more details presented in Table~\ref{table2}.

To better examine the dataset's features, we analyzed the distribution of question and answer lengths of the ViSignVQA dataset. We also looked at how many things were in each picture, paying special attention to how many signboards had to be identified in order to respond to the questions. Additionally, we analyzed the different kinds of questions, focusing on those that dealt with color in the context. Due to resource constraints, all analyses were conducted solely on the development set.  A detailed statistical overview is provided in Section \ref{Question Analysis} and Section \ref{Answer Analysis}.
\subsubsection{Question Characteristics}
\label{Question Analysis}
\subsubsection*{Question Length}

\begin{figure}[htp]
    \centering
    \includegraphics[width=1\textwidth]{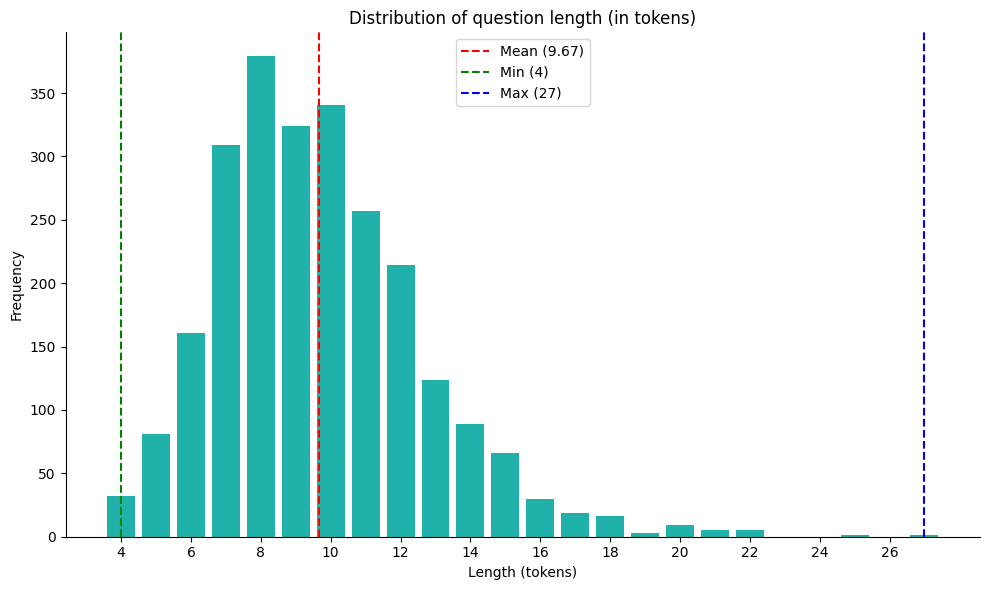}
    \caption{Distribution of question length.}
    \label{ques_len}
\end{figure}

A statistical analysis of question length in the development set (Figure \ref{ques_len}) reveals substantial variation, ranging from 4 to 27 tokens. Short questions are common and typically address straightforward signboard information (e.g., “Where does this signboard lead to?”), while longer questions often involve more complex reasoning across multiple signboards or broader contexts. The average length of 9.59 tokens indicates a balanced distribution, ensuring that the dataset captures both simple and challenging question types.

\subsubsection*{POS Tag Analysis of Questions}
\begin{figure*}[htp]
  \centering
  \begin{minipage}{0.58\textwidth}
    \centering
    \includegraphics[width=\textwidth,height=4.5cm]{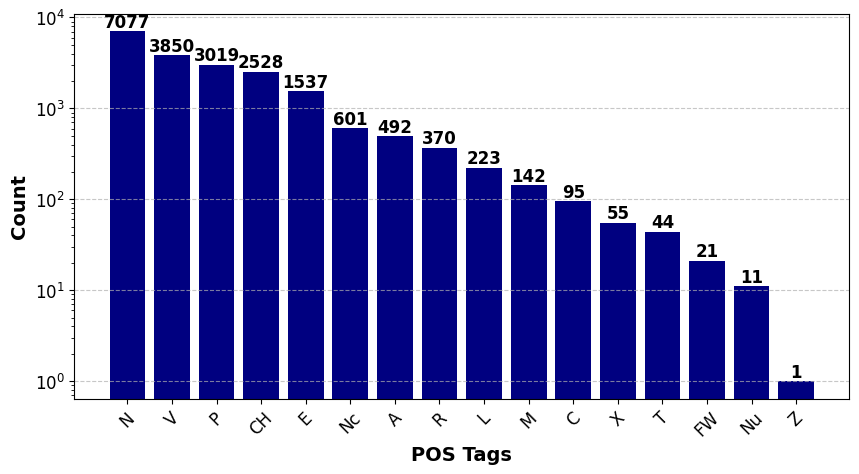}
    \caption{Distribution of Postag in question.}
    \label{pos_tag_quest}
  \end{minipage}
  \hfill
  \begin{minipage}{0.38\textwidth}
    \centering
    \includegraphics[width=\textwidth,height=5cm]{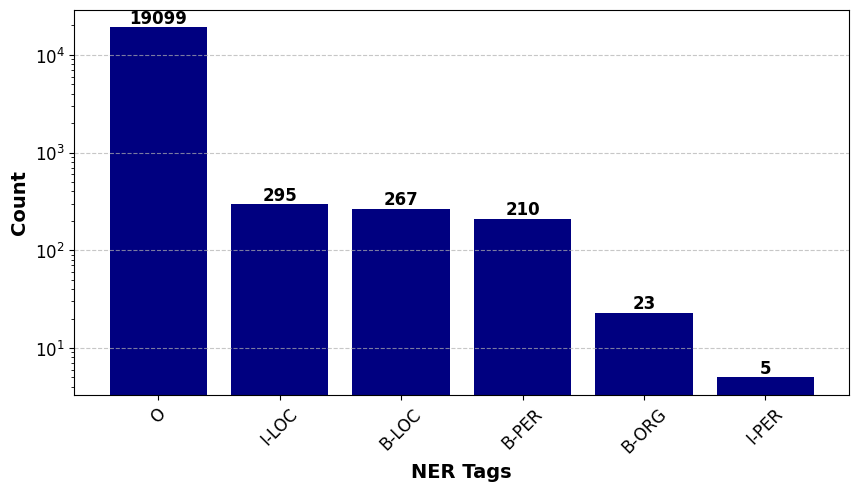}
    \caption{Distribution of NER in question.}
    \label{ner_quest}
  \end{minipage}
\end{figure*}

In the context of VQA, examining the Part-of-Speech (POS) distribution of words in questions is an important step toward understanding their syntactic structure and semantic roles. As illustrated in Figure~\ref{pos_tag_quest}, our validation set exhibits a diverse range of POS categories. Nouns (N) are the most dominant, with over 7,077 occurrences, typically denoting objects such as signboards depicted in the images. Pronouns (P), which occur more than 3,019 times, serve as substitutes for nouns and play a key role in linking different elements within the question. Verbs (V) appear approximately 3,850 times, often expressing the action or condition of an object. Prepositions (E), with around 1,537 instances, occur less frequently and are mainly used to encode spatial, temporal, or contextual relations.

The tag CH also accounts for a significant portion, with 2,528 occurrences, representing punctuation marks. This indicates that many questions contain multiple clauses, suggesting a diverse syntactic structure and a higher level of complexity in question formulation.

Adjectives (A) and classifier nouns (Nc) occur with moderate frequency, each appearing in more than 400 instances. In Vietnamese, Nc denotes nouns carrying classifier functions, whereas adjectives describe specific attributes or properties. Adverbs (R) are relatively rare, with around 370 occurrences, but they play an important role in modifying actions, events, or qualities by indicating manner, degree, time, or frequency, thereby enriching the expressiveness of questions.

In addition, other categories such as numerals/quantifiers (M), determiners (L), and subordinating conjunctions (C), though less frequent, are essential for specifying and contextualizing image-related details. Despite comprising a smaller fraction of the overall text, these elements introduce syntactic diversity and grammatical complexity, posing further challenges for VQA models in handling the full range of linguistic variations.

\subsubsection*{NER Tag Analysis of Questions}

Similar to POS tagging, Named Entity Recognition (NER) offers valuable insights into the linguistic patterns of questions, with a focus on identifying entity types such as locations, organizations, persons, and miscellaneous categories. As illustrated in Figure \ref{ner_quest}, the majority of tokens in the development set are labeled as “O”, with over 19,000 occurrences. This category corresponds to words that are not recognized as named entities, typically consisting of common vocabulary describing objects, people, or actions relevant to the signboard context.

Additionally, specific named entities, such as locations (I-LOC, B-LOC), appear relatively infrequently, with 295 and 267 occurrences, respectively. Other entity labels, such as Person (B-PER, I-PER) and Organization (B-ORG), also contribute only a small proportion to the dataset.

This discrepancy can be attributed to the fact that questions in the dataset tend to be concise and typically focus on extracting specific textual information rather than providing broader contextual details. In particular, location-related contexts of signboards are referenced less frequently, as the primary goal of the questions is to retrieve key text elements rather than describe geographic or organizational aspects in detail.

\subsubsection*{Distribution of Wh-Question Types}
Wh-questions form a crucial component of the visual question-answering (VQA) task, as they target specific information related to the images. Understanding the distribution of these question types provides valuable insights into the types of inquiries our dataset encourages, which ultimately influences the models' ability to respond effectively.We extracted Wh-Question Types following these steps:
\begin{itemize}
    \item \textbf{1: }We used deep-translator (a Python package) to translate questions from Vietnamese to English.
    \item \textbf{2: }We wrote code to detect question words such as What, Where, How, etc. for each question.
    \item \textbf{3: }We manually checked whether the machine detected the question types correctly.
\end{itemize}
Finally, we compiled the data.

\begin{figure}[H]
    \centering
    \includegraphics[width=1\textwidth]{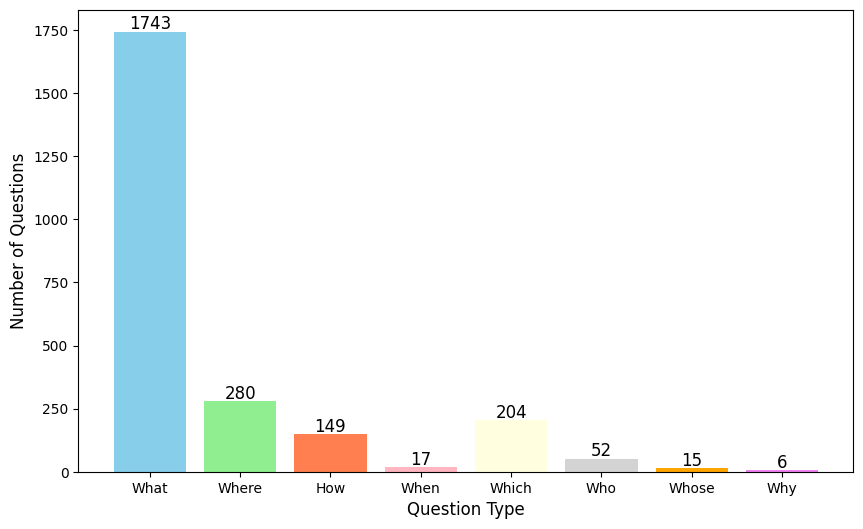}
    \caption{Distribution of Wh-Question Types in ViSignVQA Dataset (Development Set)}
    \label{wh_distribution}
\end{figure}

As can be seen in Figure \ref{wh_distribution}, What questions are the most prevalent. These questions primarily seek to identify or describe objects, events, or textual content displayed on the signboards. The dominance of "What" questions reflects the dataset's strong focus on extracting specific information from signboards, such as names, numbers, or general descriptions

Where questions are the second most common, reflecting inquiries about the location of objects or text on the signboards. This suggests that spatial awareness and understanding of signboard placement are key for answering such questions.

Which questions also appear frequently and typically ask the model to select between multiple options. These questions often require distinguishing between objects, which can involve recognizing visual details like color, shape, or relative positioning.

Less frequent question types include How, Who, Whose, When, and Why. These types of questions often require contextual reasoning or knowledge beyond basic object identification, adding complexity to the VQA task. For instance, How questions might ask about methods or processes depicted on signboards, while Who and Whose questions might involve identifying ownership or authorship.

\subsubsection*{Color Feature Analysis}
\begin{figure*}[h]
  \centering
  \begin{minipage}{0.4\textwidth}
    
    \includegraphics[width=4cm,height=4.5cm]{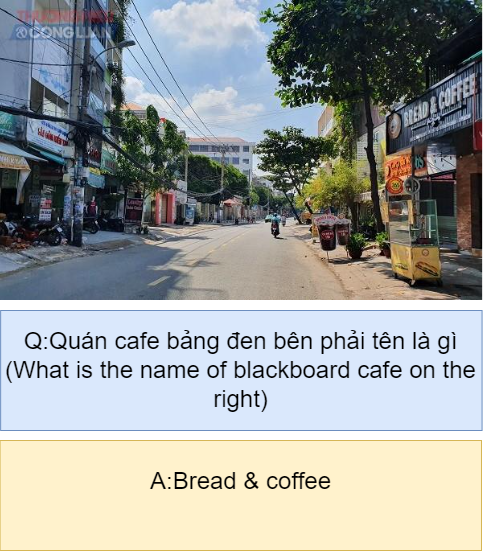}

    \label{pos_tag_ques}
  \end{minipage}
  \begin{minipage}{0.38\textwidth}
    \centering
    \includegraphics[width=4cm,height=4.5cm]{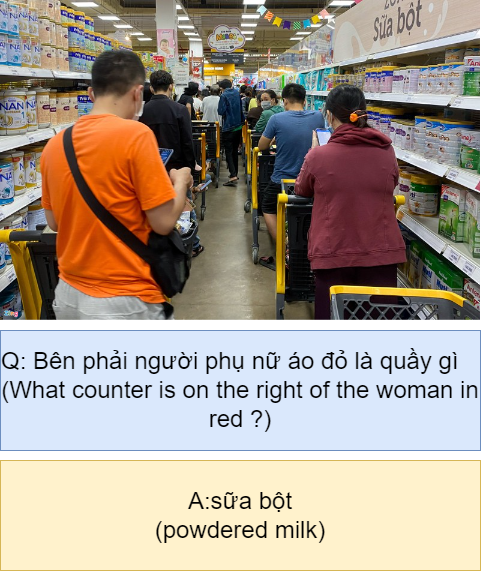}
    
    \label{ner_ques}
  \end{minipage}
  \caption{Samples from the ViSignVQA dataset require color detection to answer}
\end{figure*}
In the ViSignVQA dataset, we introduced an additional feature named \texttt{is\_color} to capture the necessary of color as a determinant in answering questions. The inclusion of the \texttt{is\_color} feature highlights the significance of utilizing visual elements such as color when interpreting Vietnamese signboards. In many practical scenarios, distinguishing between objects, text, or scene components based on color is essential. 

From the development set, 2,266 question-answer pairs were labeled with \texttt{is\_color} = 0, indicating that the questions could be answered without referencing the image's color and 200 samples were labeled with \texttt{is\_color} = 1, where color played a crucial role in formulating the correct answer. Although the number of instances requiring color recognition is relatively small compared to those that do not, this feature introduces a notable layer of complexity within the visual question answering (VQA) task. Models designed to perform well on this dataset must be capable of not only reading and understanding the text but also recognizing visual attributes like color to generate accurate responses. The \texttt{is\_color} feature thus adds a distinct challenge that requires models to fully comprehend the visual content of signboards

\subsubsection{Answer Characteristics}
\label{Answer Analysis}
\subsubsection*{Answer length}
\begin{figure}[H]
    \centering
    \includegraphics[width=1\textwidth]{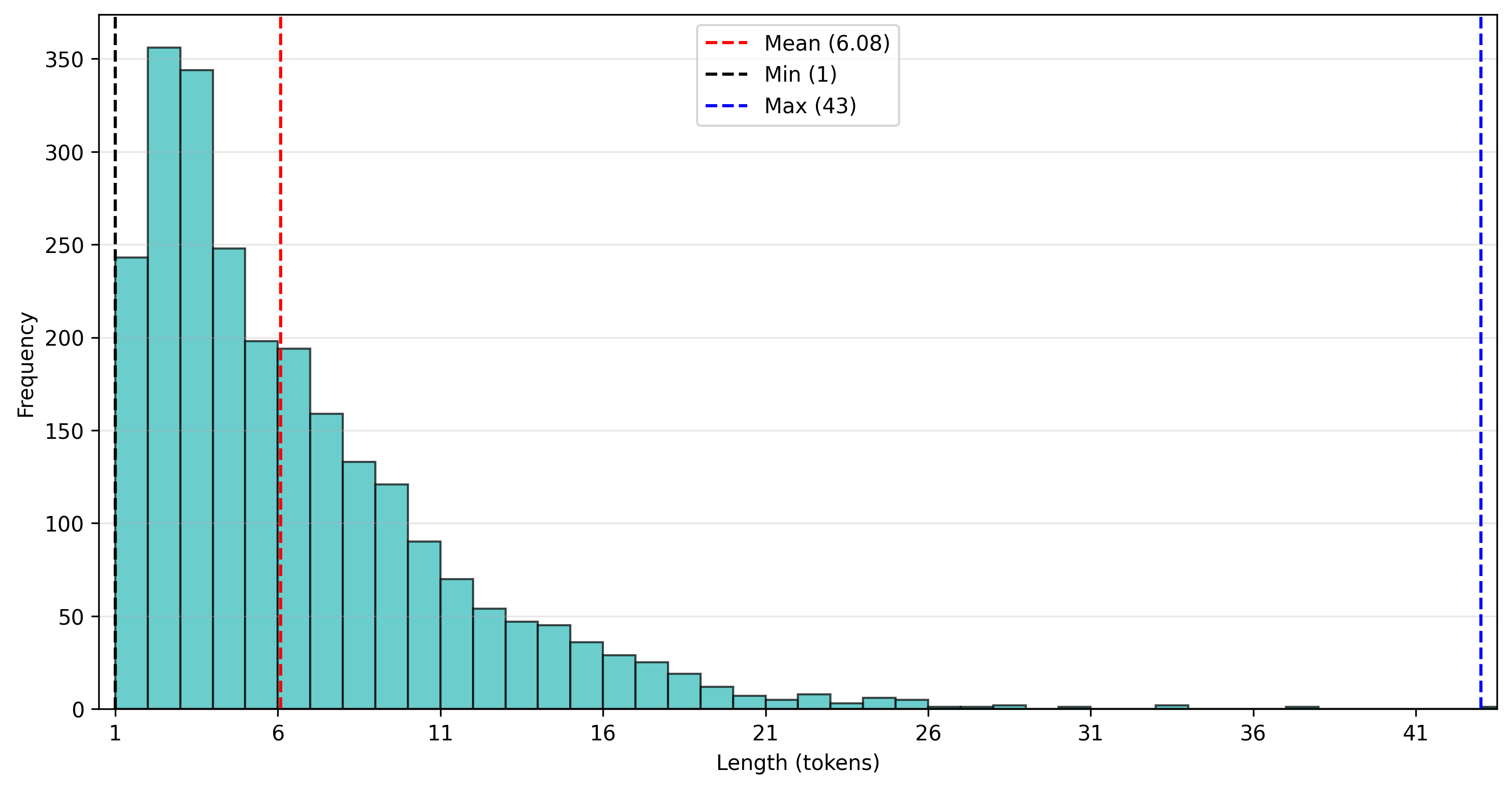}
    \caption{Distribution of answer length.}
    \label{ans_len}
\end{figure}

Figure \ref{ans_len} presents the distribution of answer lengths (measured in tokens) within our development dataset. The histogram illustrates a right-skewed distribution, indicating that shorter answers are more prevalent, while longer answers occur less frequently.

A significant concentration of answers falls within the 1-10 token range, with the highest frequency observed for answers of around 3-6 tokens. This suggests that most answers are relatively short, likely consisting of simple entity names, numbers, or concise descriptions. However, the presence of answers extending up to 43 tokens highlights cases where more detailed or descriptive responses are provided, possibly including complex phrases or multiple pieces of information.

The observed answer length distribution reflects the nature of the dataset, where short factual answers (e.g., store names, phone numbers) are dominant, while more elaborate responses, though less common, contribute to dataset diversity. This variation in answer length can pose challenges for Visual Question Answering (VQA) models, requiring them to handle both short and structured responses effectively.
\subsubsection*{POS Tag Analysis of Answer}

\begin{figure*}[htp]
  \centering
  \begin{minipage}{0.58\textwidth}
    \centering
    \includegraphics[width=\textwidth,height=4.5cm]{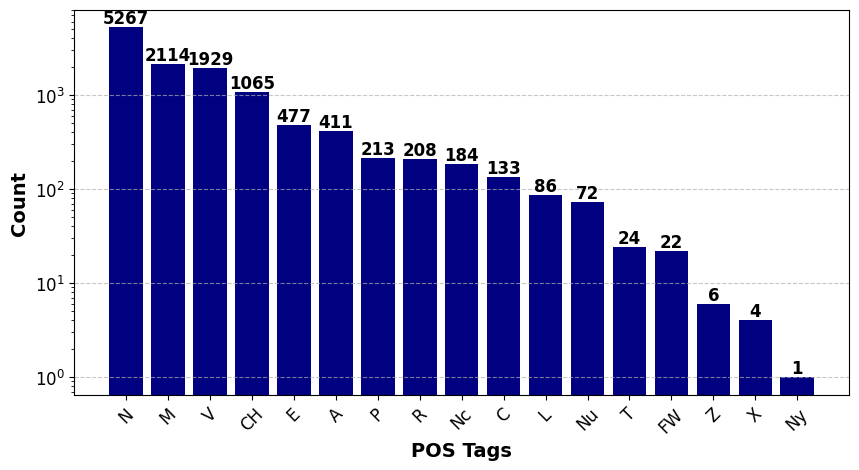}
    \caption{Distribution of Postag in answer.}
    \label{pos_tag_ans}
  \end{minipage}
  \hfill
  \begin{minipage}{0.38\textwidth}
    \centering
    \includegraphics[width=\textwidth,height=5cm]{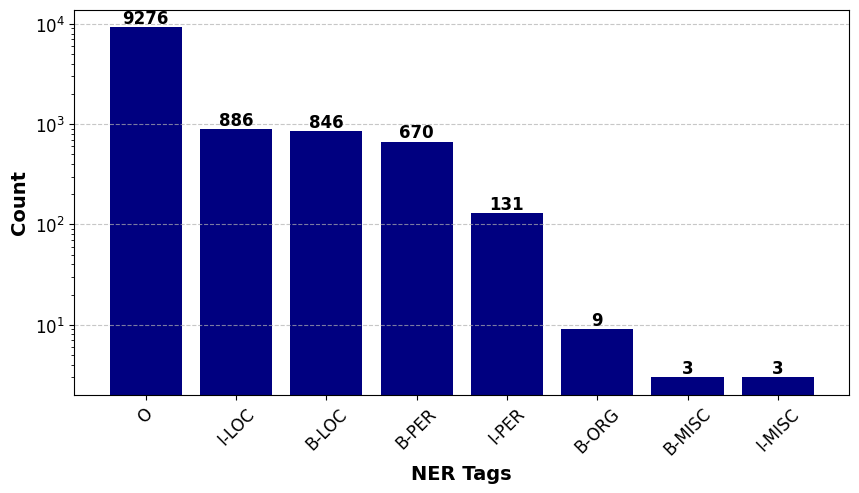}
    \caption{Distribution of NER in answer.}
    \label{ner_ans}
  \end{minipage}
\end{figure*}

As with questions, analyzing part-of-speech (POS) tags in answers provides valuable insights into both grammatical structure and semantic content, which are central to understanding how responses are constructed in VQA tasks. As shown in Figure~\ref{pos_tag_ans}, the answers exhibit notable linguistic variety. Nouns (N) account for the largest share, with about 5,200 occurrences, most often referring to objects, individuals, or other entities depicted in the images. This highlights a strong emphasis on conveying information about signboards, such as store names, phone numbers, and the surrounding context, including the presence of people near the signboards.

In addition, numerals (M) also play a significant role. With over 2,110 occurrences, ranking second after nouns (N), they indicate the diversity of answers in terms of quantity—particularly the number of signboards or people. Moreover, since answers are often based on text found on signboards, which frequently contain phone numbers (e.g., 0989.003.003) and prices, the presence of numerals is notably high, further enriching the detail and vividness of the responses.

Verbs (V) rank third in frequency, appearing approximately 1,929 times, indicating that answers often focus on human actions around the signboards, contributing to a diverse contextual representation. Additionally, the tag CH, which represents punctuation marks, follows with 1,065 occurrences. This suggests that many answers contain multiple clauses or sentence combinations, reflecting the complexity of the dataset.

Adjectives (A) are often used to describe the characteristics or conditions of the objects or entities mentioned in the answers, while prepositions (E) indicate relationships of location, time, and other contextual elements. These two categories appear with similar frequencies, at 411 and 477 occurrences, respectively. This suggests that the answers provide descriptive details about the surrounding context and objects, such as colors, locations, and other environmental attributes.

Furthermore, although less frequent, several other tags play an important role in enhancing the diversity of answers in the dataset.

\subsubsection*{NER Tag Analysis of Answer}
In addition to POS tags, NER tags are crucial for capturing specific entities in the answers. They reveal the range of entities highlighted in responses to image-based questions and provide signals that can help models attend to relevant details.

 As shown in Figure \ref{ner_ans}, the majority of words in the answers are non-entity words ("O"), with 9,276 occurrences, similar to the question dataset. However, specific named entities also appear, enriching the dataset’s diversity. Locations (B-LOC, I-LOC) are the most frequent named entities, with 846 and 886 instances, respectively, suggesting that place names often appear in signboard-related answers. Persons (B-PER, I-PER) occur 670 and 131 times, likely referencing individuals' names or business owners. Organizations (B-ORG) are much less common, with only 9 instances, indicating that company names are not a primary focus. Miscellaneous entities (B-MISC, I-MISC) are rare, appearing just 3 times each, reinforcing that the dataset primarily deals with location- and person-related information. This distribution suggests that while most answers do not explicitly mention named entities, those that do tend to focus on locations and individuals. The presence of named entities enhances the dataset’s complexity, helping OCR-VQA models better recognize and interpret structured text from images, particularly in extracting signboard information.

\subsubsection{Object Characteristics}
\label{obj_analysis}
\subsubsection*{Object in Image}
By leveraging YOLOv10 \cite{wang2024yolov10}, we obtained detailed object information from the images, including spatial coordinates, object categories, and attributes. These annotations enhance the semantic understanding of the visual content.
\begin{figure*}[htp]
    \centering
\includegraphics[width=1\textwidth]{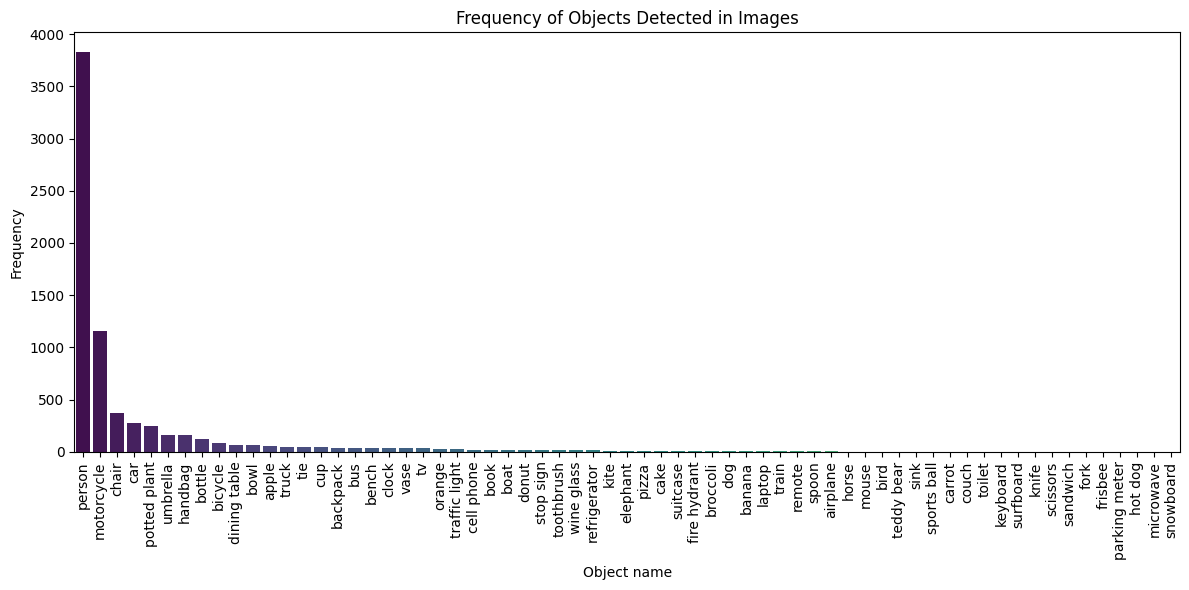}
    \caption{Distribution of the objects in images.}
    \label{fre_obj}
\end{figure*}

Figure \ref{fre_obj} The bar chart visualizes the frequency of various objects detected in a dataset, with a strong emphasis on the category "person," which appears nearly 4000 times—far exceeding the occurrences of other objects. This dominance suggests that the dataset is heavily centered around human activity, reflecting real-world image collections where humans are frequently captured.It can be observed that humans constitute a central and significant component of the images in our dataset.

Other high-frequency objects include "motorcycle" and "bicycle," which occur around 1000 times, likely representing an urban or transportation-focused dataset.Given that street signs are commonly found in urban environments, their frequent appearance is unsurprising.

Beyond these top categories, the chart displays a long tail of less frequently detected objects, such as "umbrella," "dog," and "car," all of which appear in smaller numbers. This skewed distribution, where a few categories dominate while many others appear rarely, is typical of real-world object detection tasks, highlighting the challenge of ensuring balanced representation across all object types in diverse datasets.

Additionally, the long tail of low-frequency objects, such as "microwave" or "snowboard," suggests that while the dataset includes a wide variety of items, many objects are underrepresented. This disparity can have implications for machine learning models, which tend to perform less effectively on classes with fewer examples, thereby increasing the challenge for these models to accurately recognize objects in the vicinity of street signs when answering related queries.

An important cultural insight captured in our dataset is the predominance of motorbikes over cars. As shown in Figure \ref{fre_obj}, motorbikes appear with significantly greater frequency than cars. With nearly 9,000 instances, motorbikes appear almost three times as often as cars, which are detected just over 3,000 times. This stark contrast emphasizes the prevalent use of motorbikes in daily life and the transportation system in Vietnam, offering insight into the evolving cultural trends and lifestyle choices of the population.
\subsubsection*{Signboard in Image}
\begin{figure*}[htp]
    \centering
\includegraphics[width=1\textwidth]{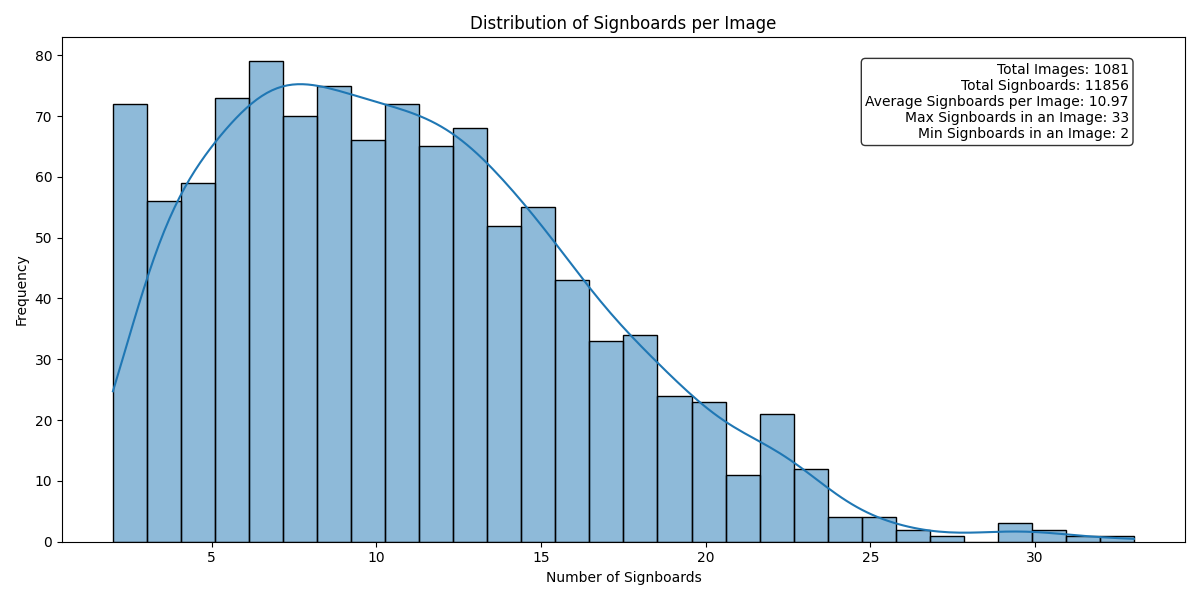}
    \caption{Distribution of the signboard in images.}
    \label{fre_sgn}
\end{figure*}
Figure \ref{fre_sgn} illustrates the distribution of signboards present within the dataset's images. The dataset contains 1,081 images and a total of 11,856 signboards, leading to an average of 10.97 signboards per image. The graph provides a detailed view of how many signboards appear in each image, with the x-axis representing the number of signboards per image and the y-axis showing the frequency of occurrences for each value.

As shown in the figure, an image in the dev set contains at least 2 signboards, with some images having up to more than 30 signboards—a substantial number that requires detection to answer the associated questions. The distribution follows a right-skewed pattern, where most images contain a moderate number of signboards (ranging from 5 to 15). The frequency peaks at around 8 signboards per image, making this the most common case.

This distribution reflects the diversity of the dataset in terms of signboard density across images, offering a balanced mix of simpler scenes with fewer signboards and more complex ones. Such variety ensures that models trained on this dataset will be exposed to both straightforward and challenging visual contexts, promoting robustness in OCR and question-answering performance.

\subsubsection*{Analysis of Objects in Questions}

We utilized VNCoreNLP \cite{vu2018vncorenlp}  to extract POS tags from the questions, enabling statistical analysis of object frequencies and revealing annotator focus within the dataset. As illustrated in Figure 15, objects such as “store” and “name” appear with particularly high frequency (over 8,000 and nearly 7,000 times, respectively), underscoring annotators’ emphasis on identifying store-related elements and specific names in the images. Visual-content terms like “photo” and “image” are also common, totaling around 10,000 occurrences, which reflects a strong focus on image details. Numerical and transactional information is well represented, with “number” exceeding 4,500 occurrences and “price” over 1,500, highlighting interest in quantifiable aspects of the scene.

\begin{figure*}[htp]
    \centering
\includegraphics[width=1\textwidth]{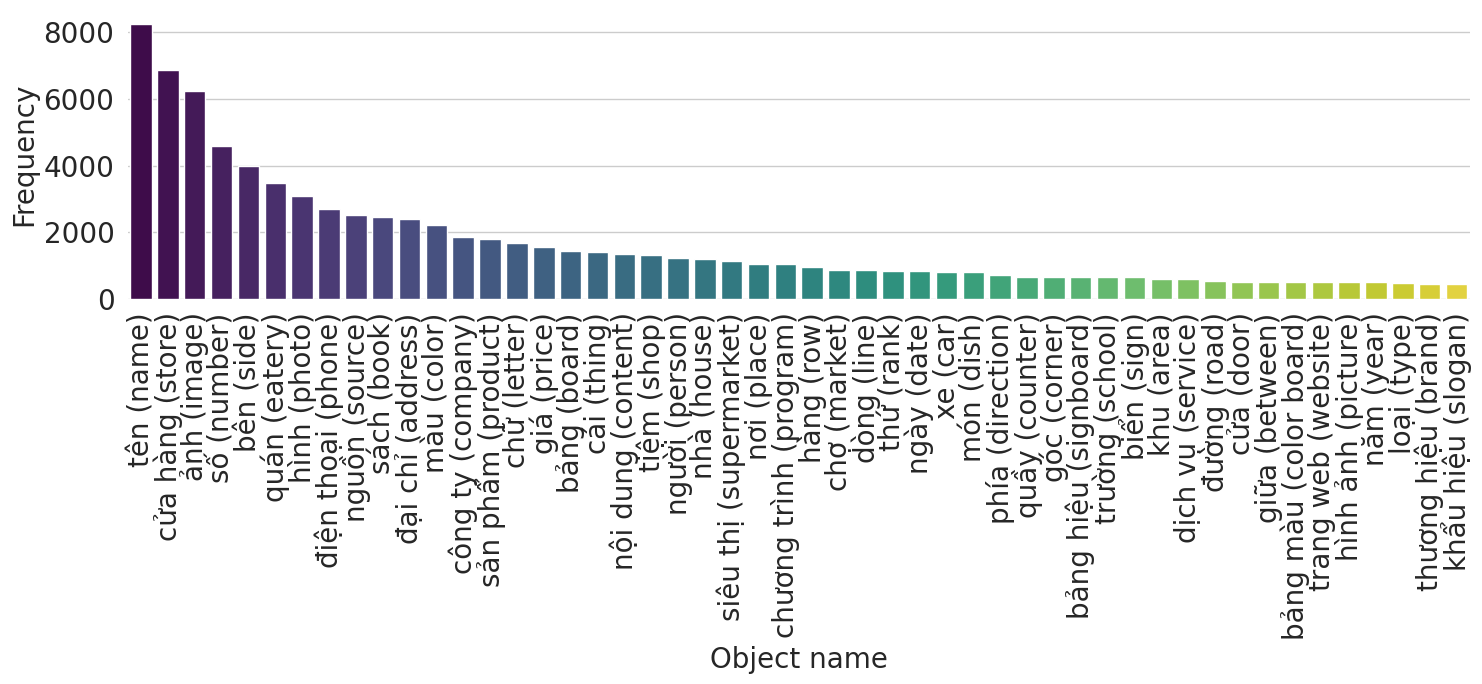}
    \caption{Top 50 Objects Referenced in Questions and Their Distribution.}
    \label{obj_ques}
\end{figure*}

Beyond this, the dataset captures cultural and contextual interests: culinary terms (“eatery”, “dish”) indicate attention to food-related content, while human-related entities (“person”, “man”, “woman”) occur at high rates, emphasizing the role of people in the visual scenes. Additionally, commercial location terms such as “shop”, “supermarket”, and “market” appear prominently, further illustrating annotators’ interest in spaces of trade and daily life.

\subsubsection{Signboard Features Analysis}
\subsubsection*{Bilingual Integration: Vietnamese and Foreign Languages}
A common characteristic of Vietnamese signboards is the blending of Vietnamese with foreign—primarily English—terms. This bilingual composition is often seen in combinations such as “Cà phê Coffee”, “Tiệm Nail Spa”, or “Bánh Mì Sandwich”. The use of English terms lends a modern and international feel, while Vietnamese words preserve cultural identity and ensure accessibility for local customers.

This strategy reflects both practical and sociolinguistic purposes. Practically, it helps businesses target a wider demographic, including tourists and younger Vietnamese who are increasingly exposed to English through media and education. Culturally, it illustrates a form of linguistic hybridization, where global and local elements coexist on a single visual surface.

Moreover, the positioning of each language on the signboard is often intentional: Vietnamese may be placed more prominently for identity, while English is used as a supporting element to signal modernity or international relevance. This bilingual integration not only facilitates communication but also reflects Vietnam’s dynamic social and cultural transformation in the era of globalization.

\begin{figure}[htp]
  \centering
\includegraphics[width=0.49\textwidth]{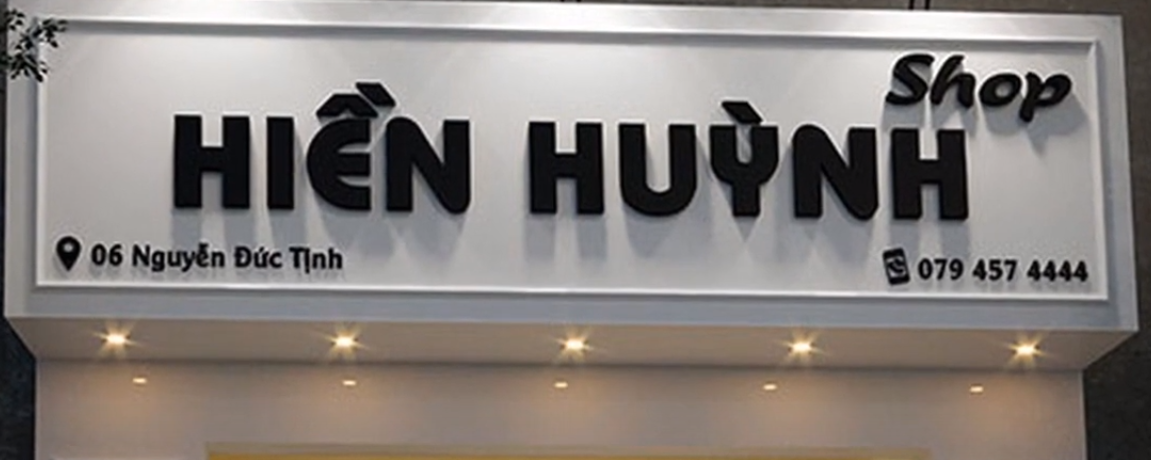}
\includegraphics[width=0.49\textwidth]{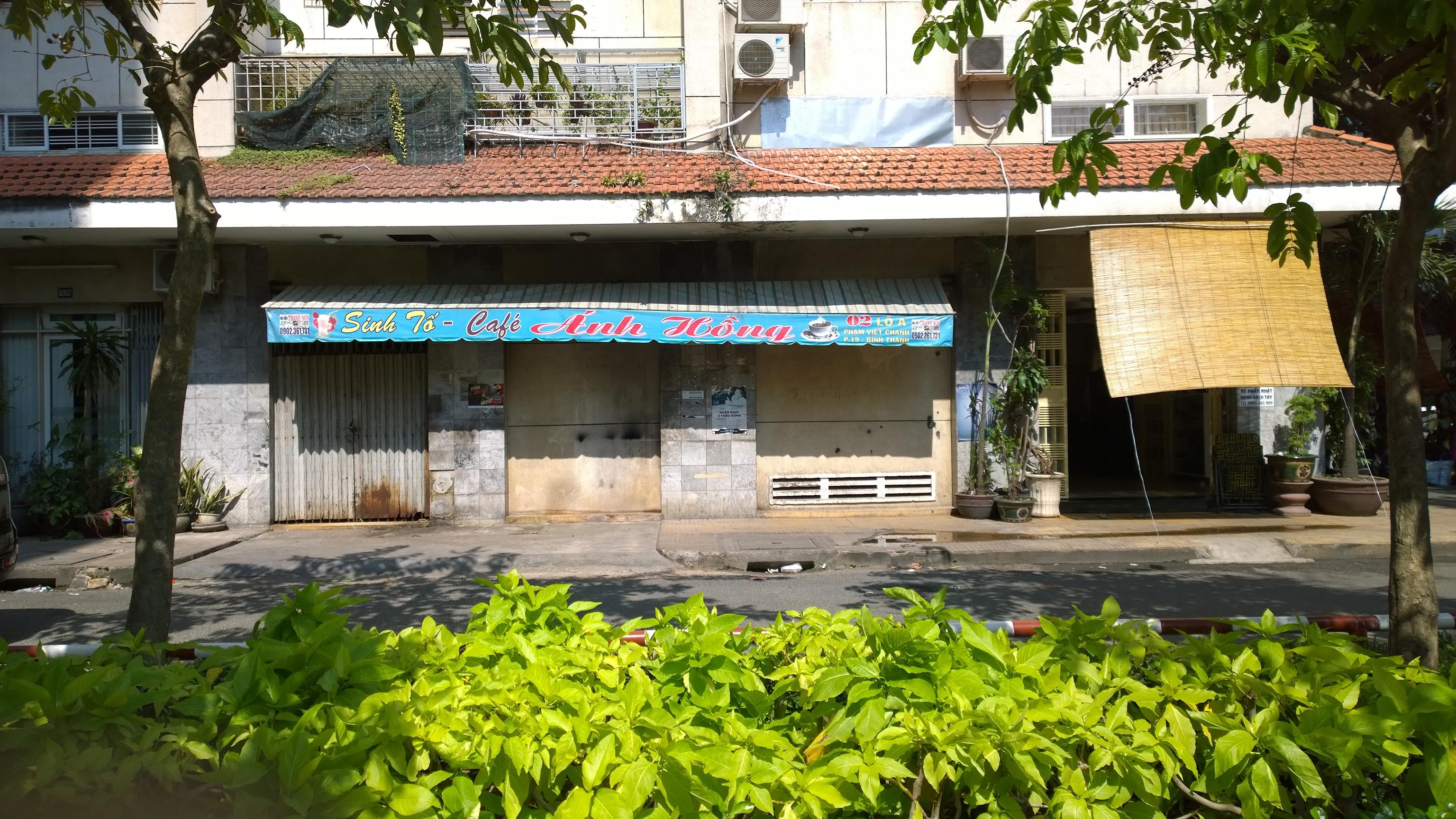}
  \caption{Example of Bilingual Integration signboard in Vietnam.}
  \label{Bilingual Integration}
\end{figure}

\subsubsection*{Concise and Succinct Language}
Vietnamese signboards prioritize brevity to convey messages instantly. Phrases are typically limited to two or three words—examples include “Bún Bò”, “Tiệm Nail”, or “Cà Phê Đắng”. Such minimal text ensures that pedestrians and motorists can read and comprehend the offer in a fraction of a second.

This economy of words serves both functional and aesthetic purposes. Functionally, it reduces visual clutter, making key information (shop type or signature product) immediately legible even at a distance or in motion. Aesthetically, the spare wording creates balanced layouts, allowing for ample negative space and giving prominence to any accompanying icons or images.

Moreover, the choice of vocabulary leans toward high-frequency, everyday terms familiar to local audiences. By avoiding complex or poetic phrasing, sign designers maintain universal comprehension across age groups and educational backgrounds. In essence, the concise language of Vietnamese signboards efficiently communicating identity and offerings with maximum clarity and minimal cognitive load.

\begin{figure}[htp]
  \centering
\includegraphics[width=0.49\textwidth]{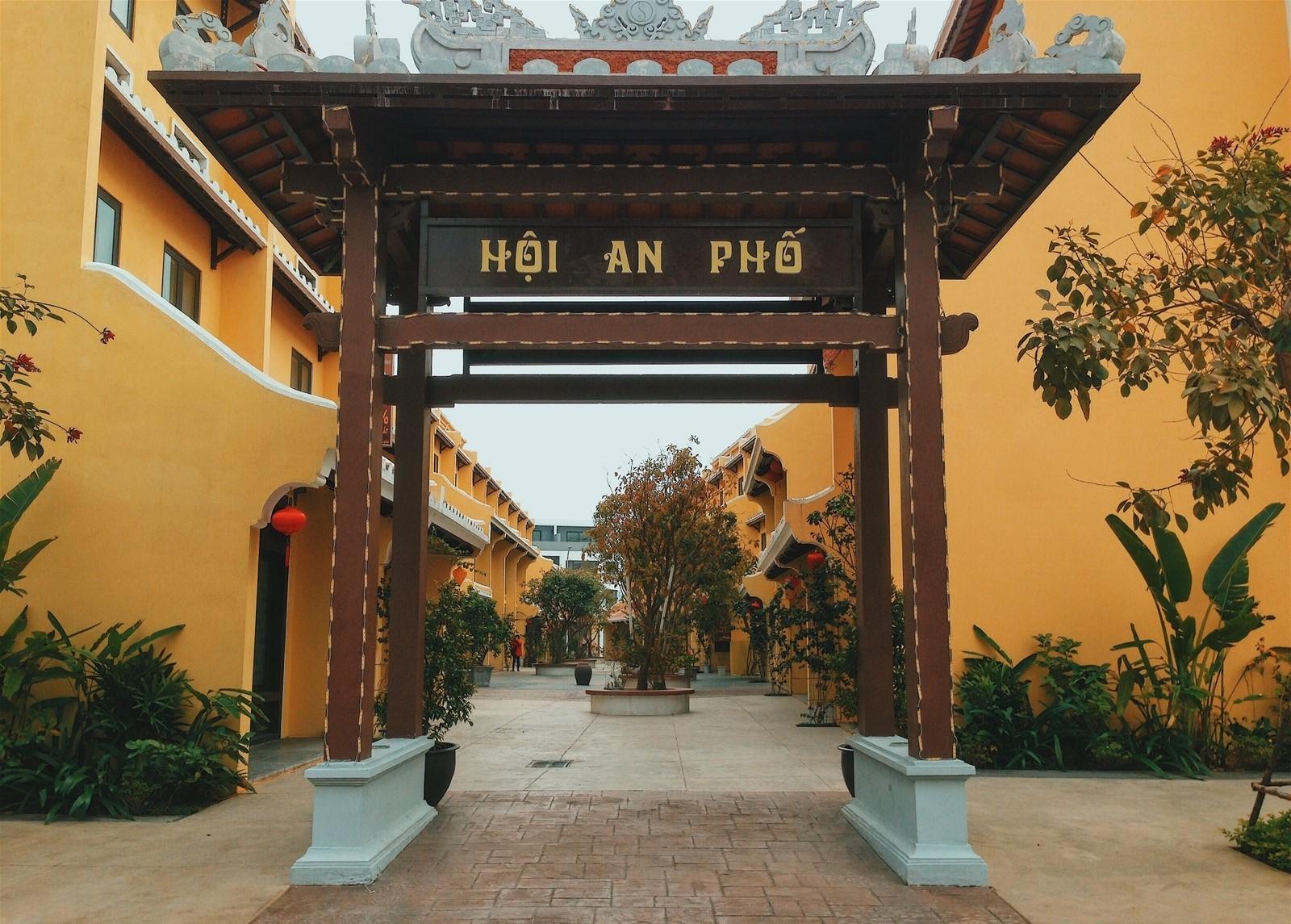}
\includegraphics[width=0.49\textwidth]{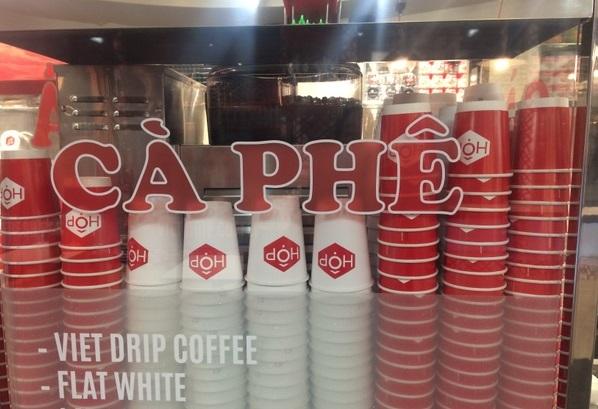}
  \caption{Example of Concise Language signboard in Vietnam.}
  \label{Concise Language}
\end{figure}

\subsubsection*{Product-Centric Imagery}
Vietnamese signboards frequently pair text with realistic or stylized depictions of signature products—steaming bowls of phở, frothy cups of cà phê, crisp bánh mì loaves—to create instant visual recognition. By positioning these images prominently (often to the left or above the shop name), designers use “visual shortcuts” that reduce cognitive effort: potential customers immediately grasp the core offering without reading a single word.

This strategy serves multiple purposes. Functionally, it captures attention in busy streetscapes where viewers have only fractions of a second to decide where to stop. A vivid icon of a bowl or cup conveys taste, texture, and ambience in ways text alone cannot. Aesthetically, product imagery balances negative space and anchors the overall layout, lending harmony between graphics and typography.

Moreover, such imagery often reflects local presentation styles: hand-drawn sketches suggest artisanal authenticity, while photo-realistic rendering implies modern comfort. In either case, combining product visuals with succinct Vietnamese or bilingual text enhances both clarity and emotional appeal, making signboards not just labels but appetizing invitations.

\begin{figure}[htp]
  \centering
\includegraphics[width=0.49\textwidth]{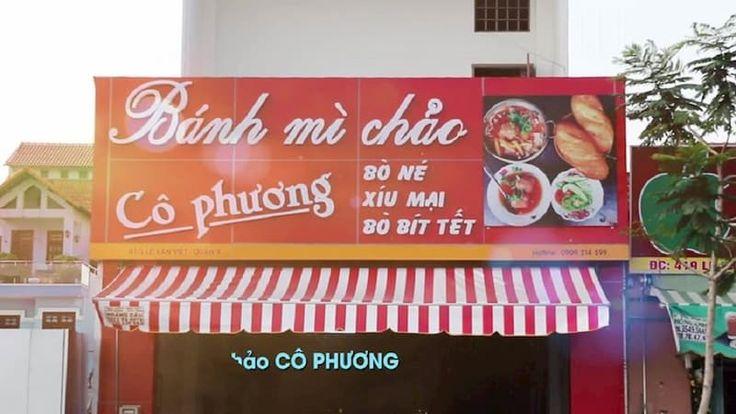}
\includegraphics[width=0.49\textwidth]{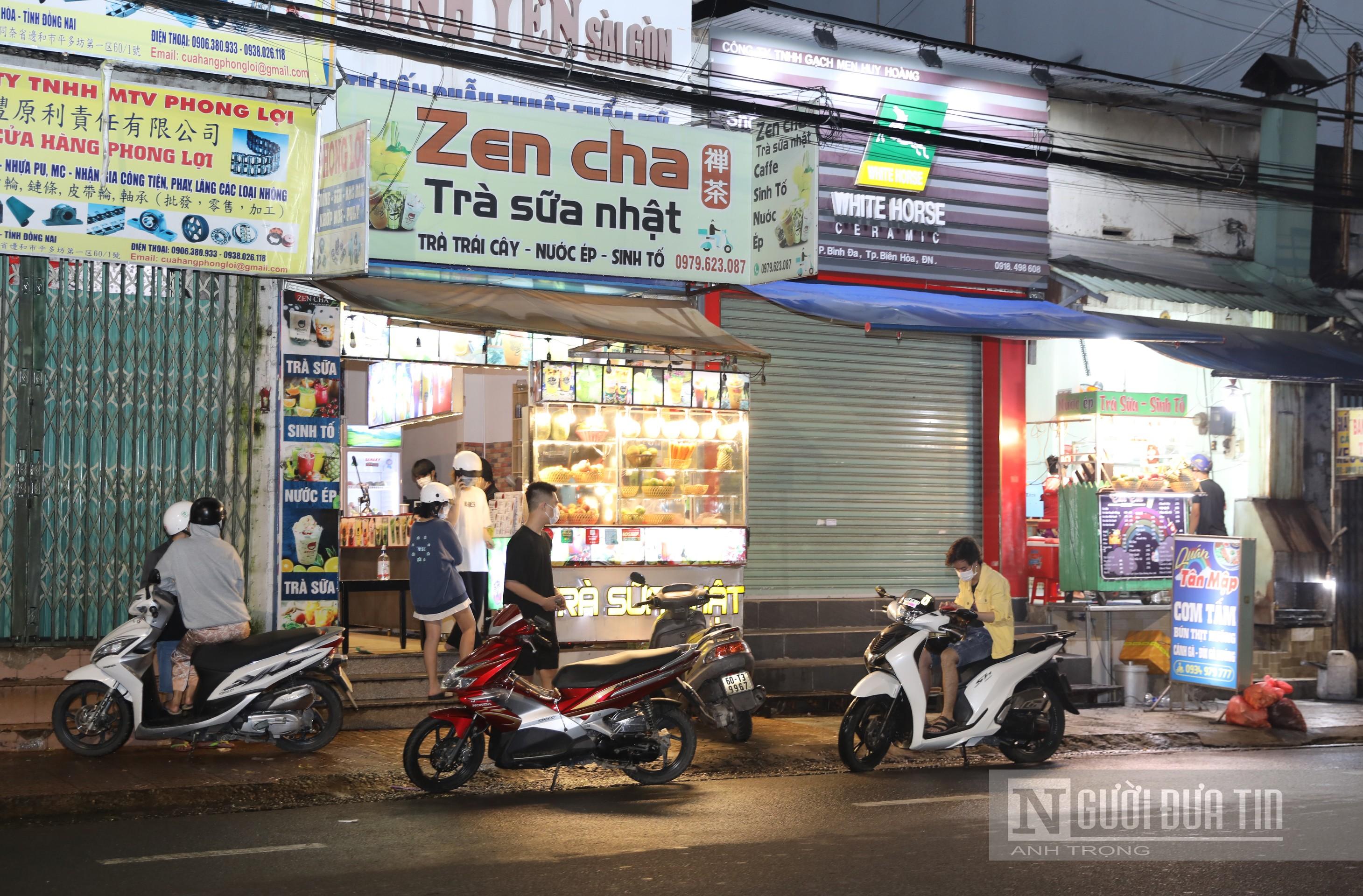}
  \caption{Example of Visual Shortcuts signboard in Vietnam.}
  \label{Visual Shortcuts}
\end{figure}

\subsubsection*{Familiar and Relatable Language}
One of the most distinctive features of Vietnamese signboards is their use of everyday, familiar language that resonates with the general public. Phrases such as “Nhà hàng” (restaurant), “Quán ăn” (eatery), or names of popular local dishes like “bún riêu,” “bánh căn,” or “hủ tiếu” are typically short, relatable, and immediately accessible. Rather than adopting ornate or formal language, Vietnamese signage is designed to reflect the way people naturally speak in daily life.

Instead of using formal descriptors like “Vietnamese traditional cuisine” or “local specialty dishes,” shop owners often opt for simpler, more evocative expressions such as “bún mắm miền Tây” (Western-style fermented noodle soup), “bánh mì chả lụa” (Vietnamese sandwich with pork roll), or “cà phê cóc” (street-style coffee). This approach not only saves space on the signboard but also allows readers to instantly visualize the item or service being offered, triggering a sense of familiarity and trust.

Additionally, descriptive adjectives related to product or service quality are often used in a friendly, conversational tone—words like “ngon” (delicious), “rẻ” (cheap), “tươi” (fresh), “nóng hổi” (steaming hot), or “đặc biệt” (special). For instance, phrases like “Phở ngon mỗi ngày” (Delicious pho every day) or “Trà sữa rẻ mà chất” (Cheap but quality milk tea) are not only concise and memorable but also create a welcoming impression through their casual tone. This everyday style of wording turns signage into more than just informational displays; they become personal and friendly invitations to potential customers.

From a technical perspective, this linguistic diversity—marked by colloquialisms, abbreviations, and non-standard sentence structures—poses significant challenges for optical character recognition (OCR) and natural language processing (NLP) systems. Generic models often struggle with local slang, regional terms, and informal syntax. Therefore, for high-performing systems, it is crucial to train on in-domain corpora—datasets collected directly from real-world signage that capture the full range of vernacular expressions and contextual variations.

\section{Methodology}\label{sec4}
\begin{figure*}[htp]
    \centering
\includegraphics[width=1\textwidth, height=5cm]{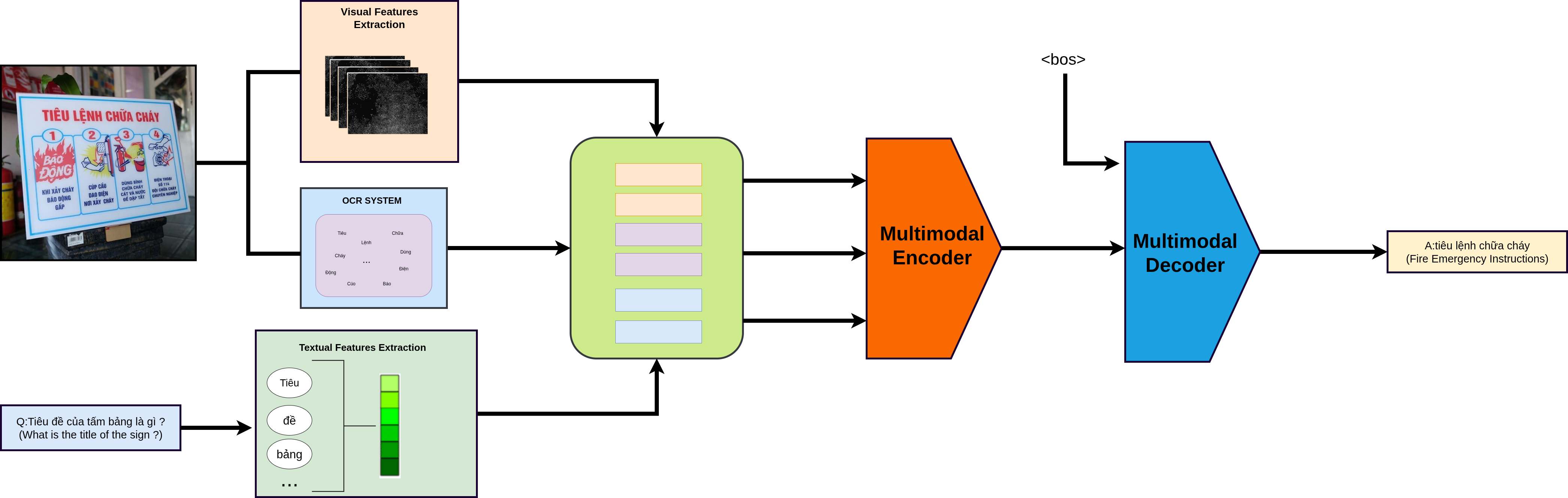}
    \caption{Structural Overview of Text-Based VQA Models.}
    \label{baseline_models_structure}
\end{figure*}

A typical text-based Visual Question Answering (VQA) framework is composed of three primary components: Visual Feature Extraction, Textual Feature Extraction, and an Optical Character Recognition (OCR) System (Figure \ref{baseline_models_structure}). The Visual Feature Extraction module captures salient visual information from the input image, while the Textual Feature Extraction module encodes the natural language question into a meaningful feature vector. In parallel, the OCR System detects and extracts textual content embedded within the image, with particular emphasis on signboards. By integrating these three components, the model is able to jointly reason over visual elements, linguistic inputs, and scene text, ultimately enhancing its capability to generate accurate and contextually relevant answers.

\subsection{Textual Features Extraction}
\label{text_pretrained}
To improve performance on the signboard VQA task and better handle the characteristics of the Vietnamese language, we incorporated pre-trained Vietnamese language models to extract textual features from both the questions and the OCR-recognized text in the baseline models. These models effectively capture the linguistic nuances and contextual meanings unique to Vietnamese, which is crucial due to the varied styles, formats, and textual content on signboards. For a detailed overview of the pre-trained models used, see Table~\ref{baseline_structure} in Section~\ref{baseline_models}. The selected models include:

ViT5\cite{phan2022vit5}: ViT5 (Vietnamese Text-to-Text Transfer Transformer), introduced in \cite{phan2022vit5}, is a state-of-the-art encoder-decoder model tailored for Vietnamese. Developed by VietAI, ViT5 is trained on a large and diverse corpus of high-quality Vietnamese texts using a T5-style self-supervised learning strategy. This enables the model to handle a variety of natural language understanding and generation tasks, such as text comprehension, question answering, and text-to-text transformation, making it well-suited for understanding both questions and OCR-derived signboard content in the Vietnamese context.

\subsection{Visual Features Extraction}
\label{vision_pretrained}
To leverage the strength of visual representations in VQA tasks, we adopted two prominent pre-trained image models—ViT and VinVL \cite{zhang2021vinvl}—for feature extraction in our baselines. In the English baseline models described in Section \ref{baseline_models}, we employed Vision Transformer (ViT) \cite{dosovitskiy2020vit}, a Transformer-based architecture pre-trained on large-scale image datasets, to extract global visual features that capture high-level semantic representations of the input images. In adapting these baselines to the Vietnamese context, we retained ViT for consistency in visual feature extraction.

However, when evaluating the CNN-LM based baseline models on the ViSignVQA dataset, we opted for VinVL \cite{zhang2021vinvl}, a stronger object-centric model built upon Faster R-CNN \cite{girshick2014rich}, to extract fine-grained object features. This choice not only aligns with prior work on CNN-based approaches but also facilitates object-level analysis in Section~\ref{obj_analysis}.

Vision Transformer (ViT) \cite{dosovitskiy2020vit}: The Vision Transformer (ViT) applies the Transformer architecture directly to images by dividing them into fixed-size patches. Each patch is linearly embedded and then passed through a standard Transformer encoder. This design enables the model to capture long-range dependencies and global contextual relationships across the entire image, without relying on convolutional operations. ViT has demonstrated competitive performance across various visual tasks and often surpasses traditional CNN-based models, especially when trained on sufficient data.

VinVL: While many vision-language models have focused on improving multimodal fusion strategies, VinVL \cite{zhang2021vinvl} emphasizes the importance of better visual representations through enhanced object detection. By extending the object and attribute categories and training on a much larger dataset, VinVL provides richer, more detailed object features. Built upon the Faster R-CNN framework, VinVL has achieved state-of-the-art results on multiple vision-language tasks, such as VQA, Image Captioning, and Visual Commonsense Reasoning (VCR). Its strong object-level representations make it particularly suitable for signboard understanding, where object granularity is often crucial.

\subsection{OCR engine}
\label{ocr_system}
Unlike commercial, closed-source English OCR systems frequently used in VQA baseline models (see Section~\ref{baseline_models}), such as Rosetta-en and Amazon OCR, SwinTextSpotter \cite{huang2022swintextspotter} provides an open-source alternative with explicit support for the Vietnamese language. This not only enhances its applicability to our task but also promotes reproducibility and research transparency, as the model can be freely accessed, adapted, and benchmarked by the research community.
This feature makes it particularly suitable for extracting OCR features tailored to Vietnamese text, which is crucial in signboard VQA tasks where text content is central. Consequently, we chose SwinTextSpotter to serve as the primary tool for pulling OCR-derived text elements into our Vietnamese-modified foundational systems. This choice empowers the systems to more effectively decode and interpret text within Vietnamese image settings, resulting in stronger task outcomes.

SwinTextSpotter: Unlike most (SOTA) state-of-the-art OCR systems that employ a shared backbone for both text detection and recognition—potentially limiting the interaction between the two tasks—SwinTextSpotter introduces a unified, end-to-end framework designed to enhance the synergy between detection and recognition components. Built upon the Swin Transformer architecture, the model achieves improved accuracy in scene text spotting across various languages, including Vietnamese. Its ability to jointly optimize detection and recognition enables more robust text extraction, especially in complex real-world scenarios like signboards that feature diverse fonts, layouts, and visual noise.
Scene text spotting plays a particularly critical role in the signboard VQA task due to the unique characteristics of signboard imagery. Unlike standard document text, text on signboards often appears in a wide variety of fonts, sizes, colors, orientations, and layouts. Moreover, many signboards include noisy backgrounds, decorative elements, or low-contrast text, which makes accurate detection and recognition more challenging. In the Vietnamese context, these challenges are further compounded by diacritical marks and compound words, which require OCR systems to be sensitive to the subtleties of the language.

SwinTextSpotter’s end-to-end framework and transformer-based architecture allow it to better handle such variations, making it well-suited for the complex visual-linguistic patterns found in real-world signboards. By using SwinTextSpotter, we ensure that our baseline models can leverage high-quality, Vietnamese-aware OCR features, which are essential for accurately understanding and answering questions related to signboard content.


\subsection{Reference Models}
\label{baseline_models}
\begin{figure*}[!ht]
    \centering
\includegraphics[width=0.7\textwidth]{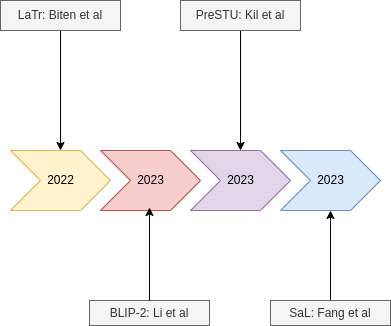}
    \caption{Timeline of Visual Question Answering Method.}
    \label{fig:timelinemodel}
\end{figure*}

A wide range of effective methods have been proposed worldwide to address the VQA task. To assess the difficulty of the ViSignVQA dataset, we selected LaTr \cite{biten2022latr}, PreSTU \cite{kil2023prestu}, BLIP-2 \cite{li2023blip}, and SaL \cite{fang2023separate} as baseline models, reflecting key milestones in the historical development of VQA methods (see Figure \ref{fig:timelinemodel}). Although these baselines were originally designed for English, we adapted them to Vietnamese while preserving their core architectures and functionalities.

\begin{table}[htp]
\caption{OCR and Feature Extraction in Baseline Models}
\label{baseline_structure}
\begin{tabular}{lcccc}
\hline
\textbf{Model} & \textbf{Visual Features} & \textbf{Textual Features} & \textbf{OCR System} & \textbf{Method Type}\\ 
\hline

BLIP-2         & ViT                   & ViT5                     & SwinTextSpotter    & ViT-LM \\
LaTr           & ViT                   & ViT5                     & SwinTextSpotter    & ViT-LM \\
SaL            & VinVL                   & ViT5                  & SwinTextSpotter     & CNN-LM \\
PreSTU         & ViT                   & ViT5                     & SwinTextSpotter    & ViT-LM \\ 
\hline
\end{tabular}
\end{table}

For an in-depth discussion of baseline approaches, we refer readers to Table \ref{baseline_structure}, which summarizes their key components. Additional details regarding the pre-trained language and vision models are provided in Section \ref{text_pretrained} and Section \ref{vision_pretrained}, respectively. The OCR system employed in our experiments is described in Section \ref{ocr_system}, while Section \ref{related_methods} offers further discussion of the methodological approaches underlying these baselines.


\textbf{LaTr:}
The Layout-Aware Transformer (LaTr) proposed by \citet{biten2022latr} adopts a Transformer-based encoder-decoder architecture, building on the T5 framework \cite{raffel2020exploring}. It comprises three primary modules:
(1) a language modeling module trained on document layouts, capturing both textual and spatial information;
(2) a spatial embedding module that incorporates positional information of OCR tokens via their bounding boxes; and
(3) a visual feature extractor based on the Vision Transformer (ViT) \cite{dosovitskiy2020vit}.
The embeddings from these three modalities are fused and passed through the encoder to produce a unified multi-modal representation, which the decoder then uses to generate output answers. LaTr is especially well-suited for tasks involving complex layouts and multi-modal text reasoning, such as signboard understanding.

\textbf{PreSTU:}
Pre-Training for Scene-Text Understanding (PreSTU) \cite{kil2023prestu} introduces a unique pre-training strategy tailored for scene-text-based tasks. Unlike other methods, PreSTU explicitly models spatial order by sorting OCR tokens from top-left to bottom-right, concatenating them using T5 separators (</s>). During training, these tokens are randomly split—one half serves as additional input, and the other as the prediction target.
A T5 model is then pre-trained on large-scale scene text image data, allowing it to learn text layout patterns and relationships. ViT \cite{dosovitskiy2020vit} is used to extract visual features throughout pretraining and finetuning. This approach helps the model understand both the textual and visual context of scene text, making it particularly relevant for signboard VQA where text layout and sequence matter.

\textbf{BLIP-2:}
BLIP-2 (Bootstrapping Language-Image Pre-training) \cite{li2023blip} is a multi-modal architecture that bridges frozen vision encoders and large language models (LLMs) using a lightweight query transformer (Q-Former). The vision encoder (e.g., ViT) and LLM (e.g., T5 or OPT) remain frozen during training, while the Q-Former is trained to align visual features with the LLM's input space.
This modular design enables efficient training while leveraging the strengths of large pre-trained components. BLIP-2 supports a wide range of vision-language tasks, including VQA, and has demonstrated strong performance across both general and OCR-related VQA benchmarks. Its flexible architecture makes it a strong candidate for multi-lingual and cross-domain VQA settings, including signboard understanding.

\textbf{SaL:}
Most existing text-based VQA methods either improve architecture design or pre-training techniques but tend to treat text in images as simple sequences, overlooking the rich spatial and semantic structures. To address this, \citet{fang2023separate} proposed Separate and Locate (SaL), a novel architecture incorporating two specialized modules:
\begin{itemize}
    \item Text Semantic Separation (TSS): identifies true semantic relations between words and filters out irrelevant text that may be spatially close but semantically unrelated.
    \item Spatial Circle Positioning (SCP): introduces a novel representation of spatial relations using circular embeddings, capturing both relative distances and directions between tokens.
By jointly modeling textual semantics and spatial layout, SaL achieves significant improvements on scene text VQA tasks such as TextVQA \cite{textvqa} and ST-VQA \cite{biten2019scene}, without requiring pre-training. Its effectiveness in handling noisy, unstructured text makes it highly applicable to signboard scenarios.
\end{itemize}

\subsection{Baseline LLM Models}
For our Visual Question Answering evaluation, we selected two state-of-the-art large language models to serve as baseline comparisons for our ViSignVQA dataset: DeepSeek-Reasoner and Gemini-1.5-Flash. These models were chosen based on their distinctive capabilities and complementary strengths in multimodal understanding and reasoning tasks

\textbf{DeepSeek-Reasoner\cite{deepseek2025r1}}
DeepSeek-Reasoner represents a specialized reasoning model developed by DeepSeek that employs Chain of Thought (CoT) processing to enhance response accuracy. The model generates detailed reasoning content before delivering final answers, making it particularly suitable for complex visual reasoning tasks. DeepSeek-R1 achieves performance comparable to OpenAI-o1 across mathematical, coding, and reasoning benchmarks, with reported accuracy rates of 79.8\% on AIME 2024 and 97.3\% on MATH-500 datasets. The model supports multimodal inputs and has been specifically designed to handle tasks requiring step-by-step logical reasoning.

\textbf{Gemini-1.5-Flash\cite{gemini2024_1_5_flash}}
Gemini-1.5-Flash was selected as our second baseline for its optimized speed and multimodal capabilities. This model is engineered for efficiency while maintaining high performance levels, with native support for text, image, audio, and video processing within single conversations. With a context window of 1 million tokens and 194 output tokens per second processing speed, Gemini-1.5-Flash provides an excellent balance between computational efficiency and performance quality. The model achieves 81\% accuracy on the MMLU benchmark while maintaining cost-effectiveness at 0.1 USD per million tokens.

In our baseline experiments, we adopt a zero-shot prompting approach for both DeepSeek-Reasoner and Gemini-1.5-Flash. Each input sample to the model consists of three key components
\begin{itemize}

\item Image: Visual content from the ViSignVQA test dataset

\item Question: Natural language queries in Vietnamese

\item System Prompt: "Bạn là 1 con bot trả lời câu hỏi dựa trên hình ảnh, câu trả lời của bạn phải là tiếng việt, trả lời đúng trọng tâm, giới hạn câu trả lời là 15 từ." (You are a bot that answers questions based on images, your answers must be in Vietnamese, answer to the point, limit your answer to 15 words.)
\end{itemize}
This standardized prompt ensures consistent evaluation conditions across both models while constraining response length to maintain focus and comparability. After that, to ensure uniformity in answer processing, normalization functions were applied, including: lowercasing text, removing punctuation, articles, and fixing whitespace.

\subsection{Multiagent VQA}
\begin{figure}[H]
    \centering
    \includegraphics[width=1\textwidth]{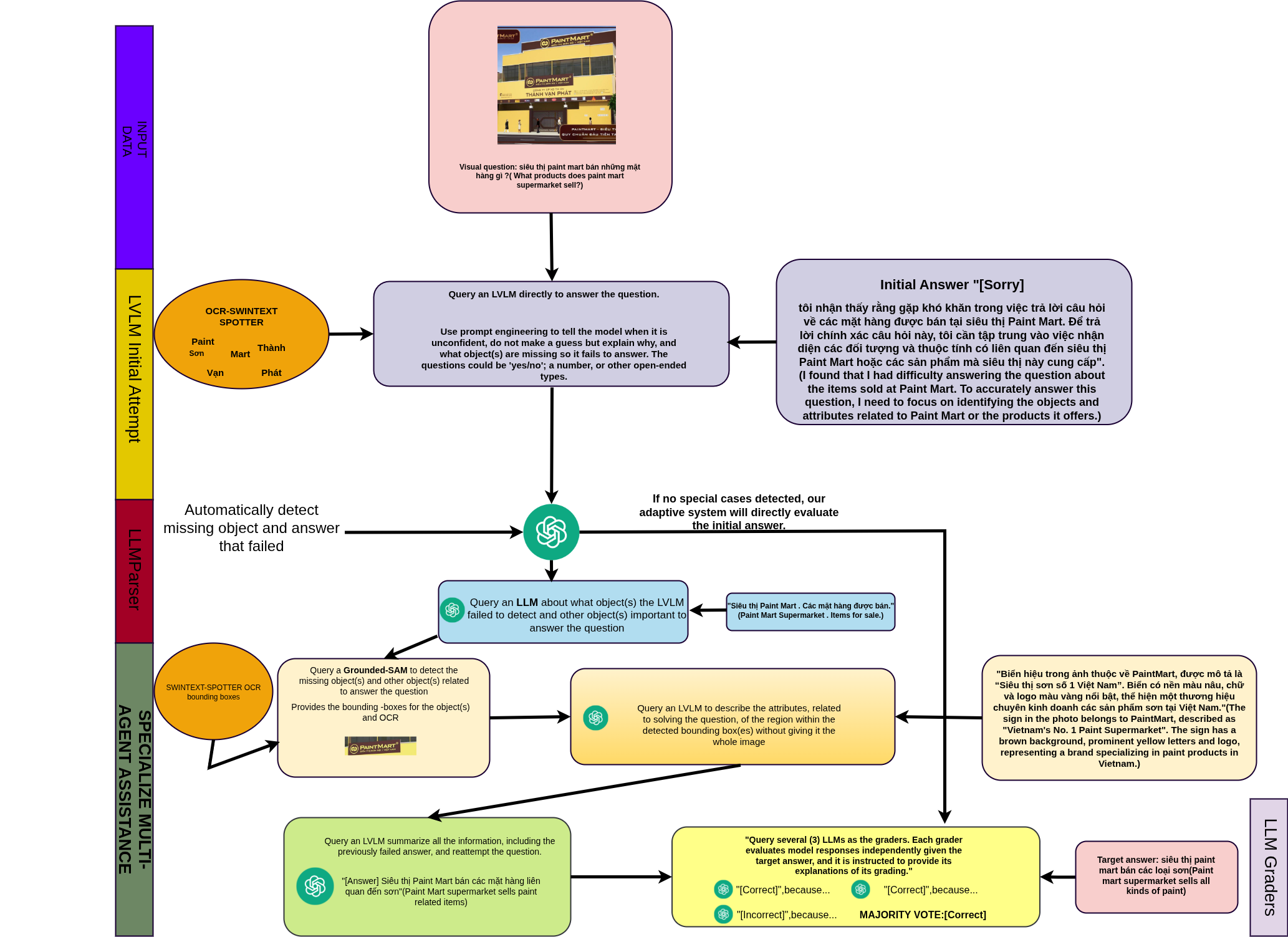}
    \caption{OURS Multi-agent Structure for Signboard-Oriented Visual Question Answering.}
    \label{OURS-agent}
\end{figure}

Based on the architecture proposed by Bowen Jiang et al\cite{jiang2024multi} , we made several modifications to better adapt it to the task of Vietnamese signboard visual question answering (VQA) (see figure ~\ref{OURS-agent}). Specifically, we introduced three key enhancements:
\begin{itemize}
    \item \textbf{Integration of an OCR module at the initial answer generation stage}: Before agents begin reasoning to produce an answer, we incorporated an OCR module(Swintextspotter) to extract textual content from the signboard. This allows the system to identify important text elements on the signboard, which are often essential for accurately answering the questions.
    \item \textbf{Utilization of OCR bounding boxes during the reattempt answer stage}: In cases where the model fails to provide a correct answer on the first attempt, we leverage the bounding box information from the OCR output in the reattempt step. This spatial information helps agents localize relevant textual regions within the image, thereby improving the alignment between the visual content and the question.
    \item \textbf{Prompt tuning tailored for the Vietnamese language}: The original prompts designed for English were carefully modified to suit the syntactic, semantic, and cultural characteristics of Vietnamese. This customization enhances the model’s ability to comprehend and respond accurately to questions in the context of Vietnamese signboards.
\end{itemize}

\subsection{Evaluation Criteria}

\subsubsection{Exact Match}
Exact Match (EM) is a widely used evaluation metric in NLP, particularly in Question Answering (QA) tasks. It measures the proportion of model predictions that exactly match the ground-truth answers, without allowing for any variation in wording or format.

In this framework, a prediction is deemed correct only if it is an exact string match with the reference answer. Although simple, EM provides a direct and interpretable measure of model performance, reflecting the frequency with which the system produces fully accurate answers.

The EM score is formally defined as:

\begin{equation}
\textbf{EM} = \frac{\text{Number of exactly matched predictions}}{\text{Total number of predictions}}
\end{equation}

\subsubsection{F1-score}
\label{sec:f1}
The F1-score is a widely used metric for evaluating system performance in Question Answering (QA) tasks, as it balances both precision and recall into a single measure. Precision reflects the proportion of predicted tokens that are correct, while recall indicates the proportion of gold standard tokens that are successfully retrieved by the system. The F1-score, defined as the harmonic mean of precision and recall, provides a more comprehensive evaluation by considering both the accuracy and completeness of the predictions.

The metrics are formally defined as follows:

\begin{equation}
\textbf{Precision} = \frac{\text{Number of correctly predicted tokens}}{\text{Total number of predicted tokens}}
\end{equation}

\begin{equation}
\textbf{Recall} = \frac{\text{Number of correctly predicted tokens}}{\text{Total number of gold standard tokens}}
\end{equation}

\begin{equation}
\textbf{F1-score} = \frac{2 \times \text{Precision} \times \text{Recall}}{\text{Precision} + \text{Recall}}
\end{equation}
\subsubsection{Accuracy-score}
Accuracy Score is a commonly used metric in classification tasks, measuring the proportion of correct predictions over the total number of instances. In our work, we adapt this metric to evaluate the performance of our multi-agent VQA system. Specifically, the accuracy reflects the number of answers deemed correct by the collective decision of agents through a majority voting mechanism.

In this setup, each agent in the system independently proposes an answer to a given question. The final prediction is determined based on majority voting among these agent-generated answers. The system is considered correct for a given instance if the majority answer matches the ground truth.

The accuracy score is formally defined as:
\begin{equation}
\mathbf{Accuracy} = \frac{N_{\text{correct}}}{N_{\text{total}}}
\end{equation}

where \( N_{\text{correct}} \) denotes the number of questions for which the majority-voted answer is correct, and \( N_{\text{total}} \) represents the total number of evaluated questions.

This metric provides a simple yet effective way to assess the collective reasoning ability and agreement level of agents within the multi-agent VQA framework.

\section{Experimental Evaluation and Results}\label{sec5}
\subsection{Experiment Settings}
\label{config}
The baseline models were trained using the Adam optimization algorithm \cite{kingma2015adam} on an NVIDIA A100 GPU equipped with 80GB of memory. Training was conducted for 10 epochs, with the ViT-LM–based method requiring approximately 7 hours to complete. The hyperparameters were configured as follows: a learning rate of $3.0 \times 10^{-5}$, a dropout rate of 0.2, a batch size of 128, and a weight decay of $1.0 \times 10^{-4}$.

For the experiments involving large language models (LLMs), we utilized the Google Colab platform and accessed the models via API calls. Each LLM experiment required approximately 8 hours of execution time. In contrast, the experiments for the multi-agent VQA system were conducted locally on our in-house infrastructure, with a total runtime of approximately 16 hours.

\subsection{Overall Results}

\begin{table}[htp]
\centering
\setlength{\tabcolsep}{10pt}
\caption{Results of  models on ViSignVQA test set.}
\begin{tabular}{lcccc}
\hline
\textbf{Model} & \textbf{Method Type} & \textbf{F1-score (\%)}     & \textbf{EM (\%)} & \textbf{Accuracy-score(\%)} \\ \hline

BLIP-2   & ViT-LM    & 40.52& 7.37&24.11\\ \hline

LaTr    & ViT-LM    & 50.3& 14.07&37.9\\ \hline

SaL     & CNN-LM    & 43.2& 9.94&22.32\\ \hline

PreSTU   & ViT-LM    & 50.37& 12.2&40.85\\ \hline
Deepseek   & LLM-base   & 17.65& 0.36&8.53 \\ \hline

Gemini-flash-1.5    & LLM-base    & 46.76& 7.09&70.48\\ \hline

GPT4     & Multiagent-LLM    & \textbf{51.76}&\textbf{18.08} & \textbf{75.98}\\ \hline

\end{tabular}
\label{Table:main_results}
\end{table}
Table~\ref{Table:main_results} reports the performance of different models on the ViSignVQA test set, evaluated by F1-score, Exact Match (EM), and Accuracy. The results highlight clear differences between vision–language models and LLM-based approaches.

For ViT-LM based methods, PreSTU achieves the highest F1-score (50.37\%), showing strong ability to produce semantically correct answers, while LaTr obtains the best EM score (14.07\%), reflecting its strength in generating exact matches. However, their Accuracy scores remain moderate (40.85\% and 37.9\%, respectively), showing that despite producing partially correct answers, exact correctness on a majority of questions is still limited. By contrast, BLIP-2 performs the weakest (Accuracy: 24.11\%), while SaL reaches a slightly higher Accuracy of 22.32\%, proving that CNN-based models remain competitive but still fall short compared to ViT-LM architectures.

For LLM-based methods, the differences are sharper. Deepseek records very low Accuracy (8.53\%), showing difficulty in handling the task because deepseek is a less trained model in Vietnamese, whereas Gemini-flash-1.5 achieves a strong Accuracy of 70.48\% alongside an F1-score of 46.76\%. This large gap demonstrates that well-trained LLMs can leverage language reasoning to make more reliable predictions, even when exact string matches (EM) remain modest.

The most striking result comes from the GPT-4 multi-agent system, which sets the overall state-of-the-art. Not only does it achieve the best F1 score (51.76\%) and EM (18.08\%), but it also obtains the highest accuracy by a large margin (75.98\%). Compared to other models, GPT-4 shows more than double the Accuracy of strong ViT-LM baselines like PreSTU, and even outperforms Gemini-flash-1.5 by over 5 percentage points. These results underline the advantage of multi-agent collaboration and majority voting, which enhance robustness and reduce individual model errors, making final predictions much more reliable.

In summary, while ViT-LM models provide solid baselines and Gemini-flash-1.5 shows the potential of LLMs, the GPT-4 multi-agent system demonstrates the most balanced and superior performance across all metrics. Its high Accuracy indicates that consensus-driven reasoning is a promising direction for future VQA research, especially in complex scenarios like Vietnamese signboards.
\section{Discussion of Results}\label{sec6}
\subsection{Analyzing the Impact of OCR-Text Suffixes in ViT5}

In our experiments, we found that models based on the ViT5 backbone outperform others when the question is enhanced with OCR text extracted from images, compared to scenarios where only the question is provided, as done in previous work. Inspired by this, we further explored how models like BLIP-2, LaTr, PreSTU, and SaL utilize OCR text when it is included as part of the question’s “context.” To achieve this, we conducted experiments by selectively including or excluding OCR text with the question, enabling a more detailed assessment of the models’ performance.

\begin{figure}[htp]
  \centering
\includegraphics[width=0.49\textwidth]{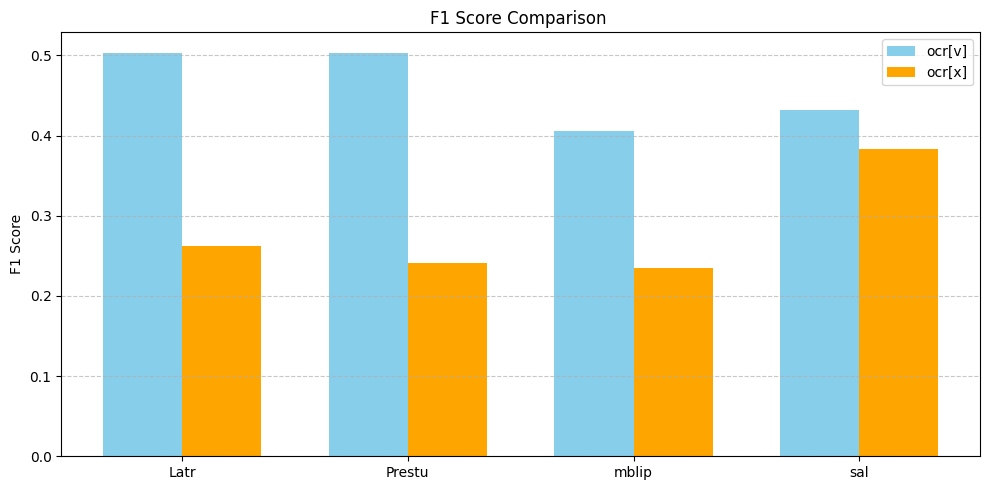}
\includegraphics[width=0.49\textwidth]{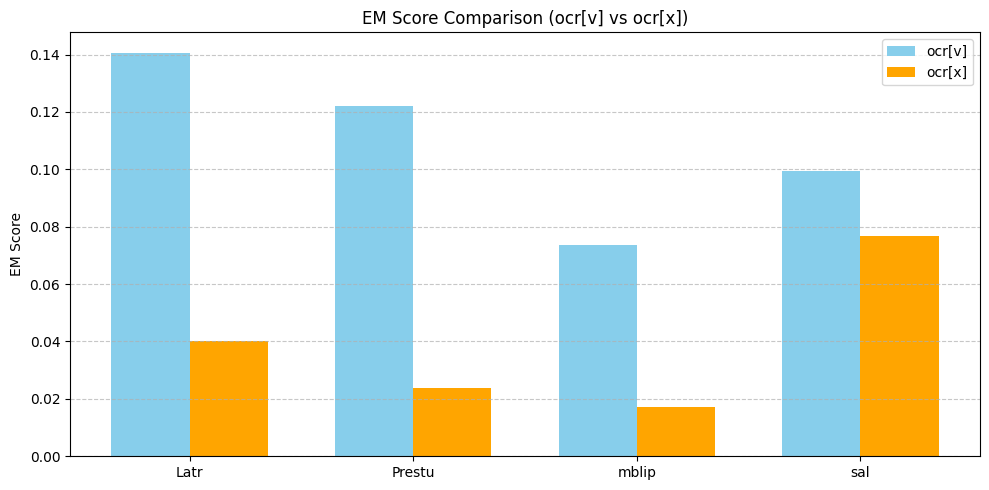}
  \caption{Performance of models include OCR text and without OCR text scenario.}
  \label{combined_figure_ocr_noocr}
\end{figure}

Figure~\ref{combined_figure_ocr_noocr} illustrates a comparative analysis of model performance with and without OCR-based context augmentation, using F1-score (left) and EM-score (right) as evaluation metrics.

Across both metrics, a consistent trend emerges: models that incorporate OCR text as part of the question input significantly outperform those that do not. This result highlights the critical role of scene text understanding in the ViSignVQA task, where answers often come directly from textual content embedded in the signboards in image.

Notably, the PreSTU model demonstrates the most substantial gain, with its F1-score increasing from 24.05\% (without OCR) to 50.37\% (with OCR), representing an improvement of nearly 209\%. Similar trends are observed in other models such as LaTr and mBLIP, further confirming the value of OCR-enhanced input. The consistent performance gap underscores the importance of textual grounding when answering visual questions involving signboards.

The EM-score comparison in the right panel of Figure~\ref{combined_figure_ocr_noocr} reinforces this observation. While all models achieve relatively modest exact match performance without OCR, adding OCR text nearly doubles the EM score in most cases. For example, LaTr's EM improves from approximately 4.0\% to over 14.0\%, while PreSTU and mBLIP show similar upward trends. These gains demonstrate that OCR context not only boosts partial matches (F1) but also facilitates precise answer selection—especially when answers are drawn directly from text regions within the image.

In summary, these findings confirm that integrating OCR text as additional context during training enables models to more effectively locate and extract relevant textual content from images, thereby enhancing their performance on Vietnamese signboard VQA tasks. This supports the intuition that many VQA answers lie in recognizing and selecting the correct span from OCR outputs, which serves as an essential bridge between visual and linguistic modalities.

\subsection{Impact of OCR Token Coverage on VQA Answerability}
\label{effect_performance}
\begin{figure}[htp]
    \centering
    \includegraphics[width=0.6\textwidth]{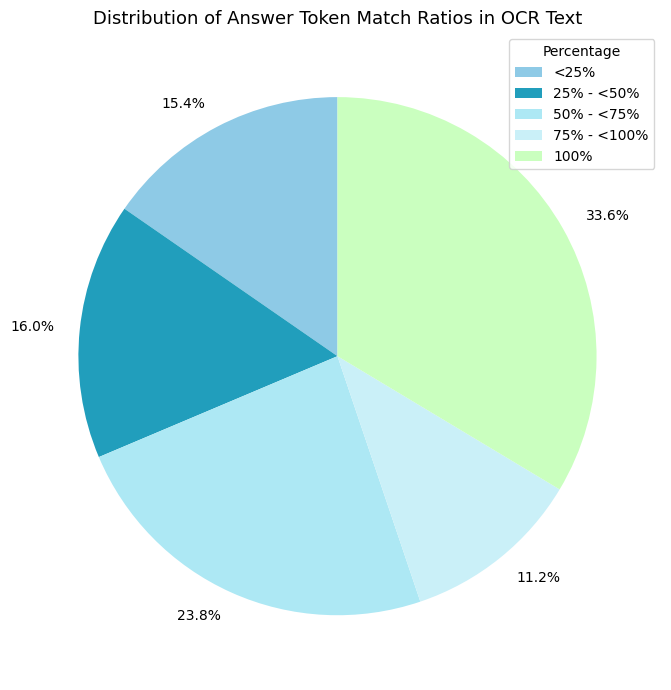}
    \caption{Answer token match ratios within OCR-extracted text across the test set.}
    \label{ocr_type_percent}
\end{figure}

Our analysis reveals a strong dependency between OCR accuracy and VQA model performance. To quantify this effect, the test set was partitioned based on the fraction of answer tokens successfully retrieved by SwinTextSpotter. Five groups were considered: <25\%, 25–50\%, 50–75\%, 75–100\%, and 100\%. Figure ~\ref{ocr_type_percent} illustrates the distribution of samples across these segments, highlighting how OCR coverage directly influences the reliability of VQA predictions.

To assess how the OCR system's performance affects the overall performance of VQA models, we performed a detailed analysis, considering both the VQA model and the frequency of answer tokens appearing in the OCR text extracted by SwinTextSpotter.
\begin{figure}[htp]
  \centering

\includegraphics[width=1\textwidth]{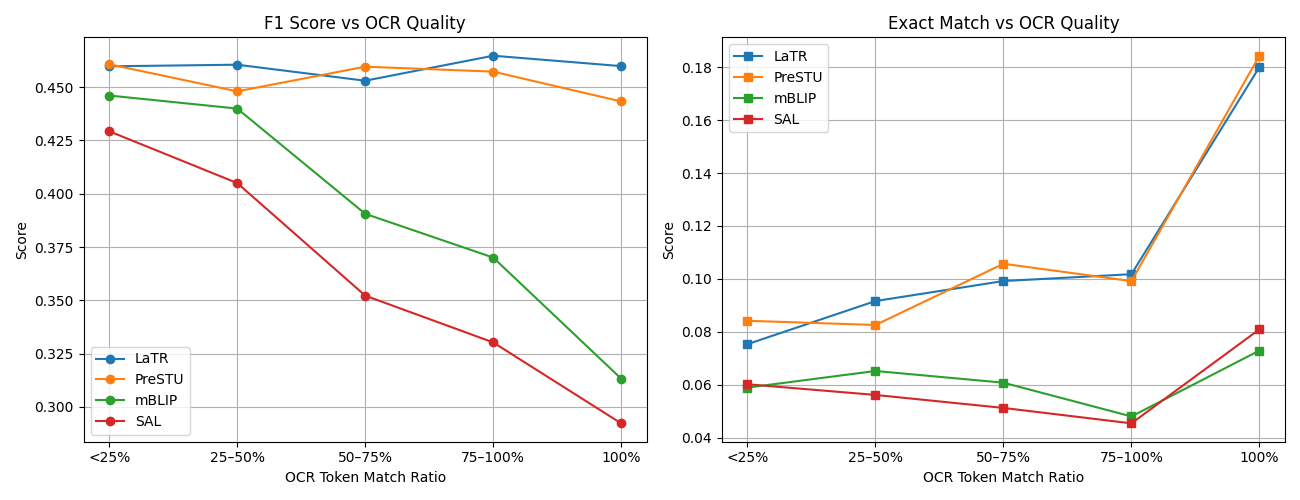}
  \caption{Effect of OCR text accuracy on the performance of LaTR, PreSTU, mBLIP, and SAL.}
  \label{combined_figure_not_sort}
\end{figure}

Figure ~\ref{combined_figure_not_sort} highlights the relationship between OCR token match ratio and model performance across four VQA baselines (LaTR, PreSTU, mBLIP, and SAL). Despite architectural differences, a consistent trend emerges among all models.

\subsubsection{F1-Score vs. OCR Quality}
As shown on the left, F1-score generally improves as the proportion of answer tokens appearing in the OCR text increases, confirming the crucial role of OCR in supporting model predictions. When OCR text contains less than 25\% of the answer tokens, model performance is significantly lower. F1 peaks at the 75–100\% range for LaTR and PreSTU, then slightly declines at 100\%, possibly due to overfitting to OCR signals or the presence of irrelevant but matching tokens.

Notably, SAL and mBLIP exhibit steeper performance degradation in low-OCR-match regions, suggesting their limited capacity to compensate for missing or noisy OCR data. Meanwhile, LaTR maintains more stable performance, indicating stronger visual-text fusion or robustness to OCR errors.

\subsubsection{Exact Match vs. OCR Quality}
The right-side chart reveals a nonlinear "zig-zag" pattern in EM scores. Performance initially declines in the 25–50\% bucket, then slightly recovers at 50–75\%, and unexpectedly dips again before rising sharply at the 100\% match ratio. This pattern suggests that having a partial match between the answer and OCR content is insufficient for exact generation, and may even mislead the model.

Importantly, even when 100\% of the answer tokens are found in the OCR text, the F1 and EM scores remain below 0.6 and 0.2 respectively, across all models. This underscores the difficulty of ViSignVQA, where OCR text alone is not a guarantee of correct answers. Effective VQA still requires robust reasoning, noise filtering, and multimodal grounding beyond surface-level token overlap.

These findings emphasize the dual challenge in OCR-VQA: models must both leverage relevant text efficiently and distinguish signal from noise in cluttered or semi-structured signboard images. Future improvements may involve adaptive OCR filtering, attention-based OCR grounding, or auxiliary supervision on OCR-to-answer alignment.
\section{Ablation study}\label{sec7}
The results of the ablation study are summarized in Tables ~\ref{tab:model_comparison_f1} and ~\ref{tab:model_comparison_em}. Overall, both OCR features and visual features contribute significantly to model performance, although their relative importance varies across architectures.

For Latr and Prestu, removing either OCR or vision features leads to substantial drops in both F1 and EM scores, highlighting their strong reliance on multimodal information. In contrast, Mblip shows almost no performance degradation when vision features are removed, while its performance drops considerably without OCR features. This indicates that Mblip predominantly exploits textual signals from OCR, with minimal benefit gained from visual cues. On the other hand, Sal exhibits the opposite trend: excluding OCR features results in only a slight decrease in performance, whereas removing vision features leads to a more noticeable reduction. This suggests that Sal relies more heavily on visual features than on OCR-derived textual information.

In summary, OCR features are crucial for Latr, Prestu, and Mblip, while visual features are more influential for Latr and Sal. These findings highlight the heterogeneous dependency patterns of different VQA models on multimodal features and suggest that the optimal feature configuration may depend on the specific model architecture.
\begin{table}[htbp]
\centering
\caption{Model Performance Comparison - F1 Score (relative to Full)}
\label{tab:model_comparison_f1}
\begin{tabular}{lcccc}
\toprule
\textbf{Configuration} & \textbf{Latr} & \textbf{Prestu} & \textbf{Mblip} & \textbf{Sal} \\
\midrule
Full           & 0.5030        & 0.5037        & 0.4052        & 0.4320        \\
No f$_{OCR}$   & 0.2624 (-0.2406) & 0.2405 (-0.2632) & 0.2354 (-0.1698) & 0.3828 (-0.0492) \\
No f$_{Vision}$ & 0.2360 (-0.2670) & 0.2360 (-0.2677) & 0.4060 (+0.0008) & 0.2320 (-0.2000) \\
\bottomrule
\end{tabular}
\end{table}
\begin{table}[htbp]
\centering
\caption{Model Performance Comparison - EM Score (relative to Full)}
\label{tab:model_comparison_em}
\begin{tabular}{lcccc}
\toprule
\textbf{Configuration} & \textbf{Latr} & \textbf{Prestu} & \textbf{Mblip} & \textbf{Sal} \\
\midrule
Full           & 0.1407        & 0.1220        & 0.0737        & 0.0994       \\
No f$_{OCR}$   & 0.0403 (-0.1004) & 0.0239 (-0.0498) & 0.0171 (-0.0823) & 0.0769 (-0.0225) \\
No f$_{Vision}$ & 0.0230 (-0.1177) & 0.0210 (-0.0527) & 0.0700 (-0.0294) & 0.0210 (-0.0784) \\
\bottomrule
\end{tabular}
\end{table}

\section{Discuss Regional Signs for the Signboard Problem Visual Question Answering}
\label{sec8}
For a language like Vietnamese, regional cultural factors are indispensable. This is especially true in the context of signboards, where regional dialects, vocabulary, and design conventions can significantly affect the performance of visual-language models.

In this study, we selected 100 images from the test set and manually labeled them according to their regional origin, including two main categories: North and South. We then filtered out the questions related to these images and evaluated the model’s performance on each regional subset.

Below, we present several observations and insights about signboards across different regions, which help explain the variation in model performance.
\subsection{Northern Signboards}
\begin{figure}[htp]
  \centering
\includegraphics[width=0.49\textwidth]{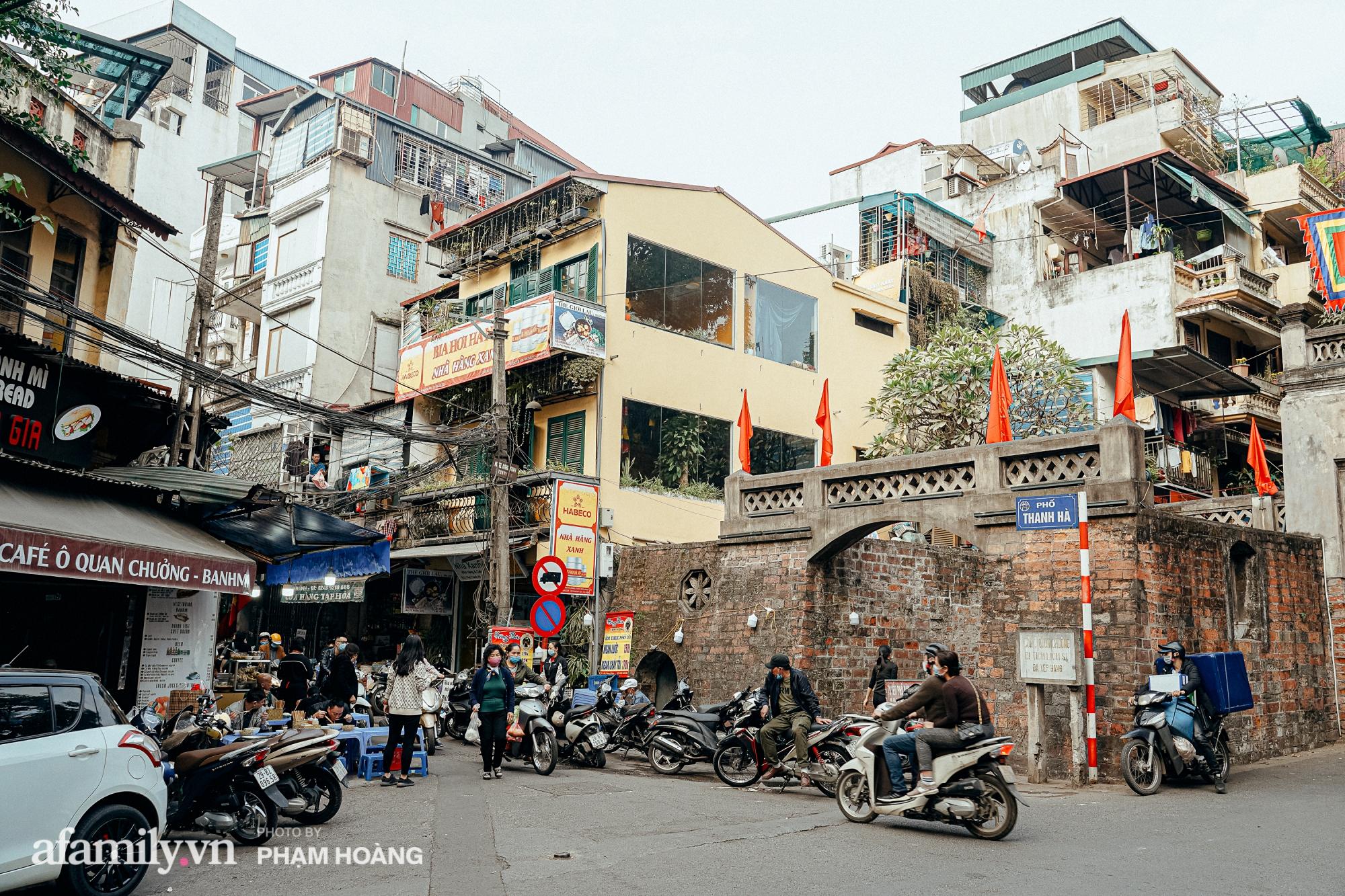}
\includegraphics[width=0.49\textwidth]{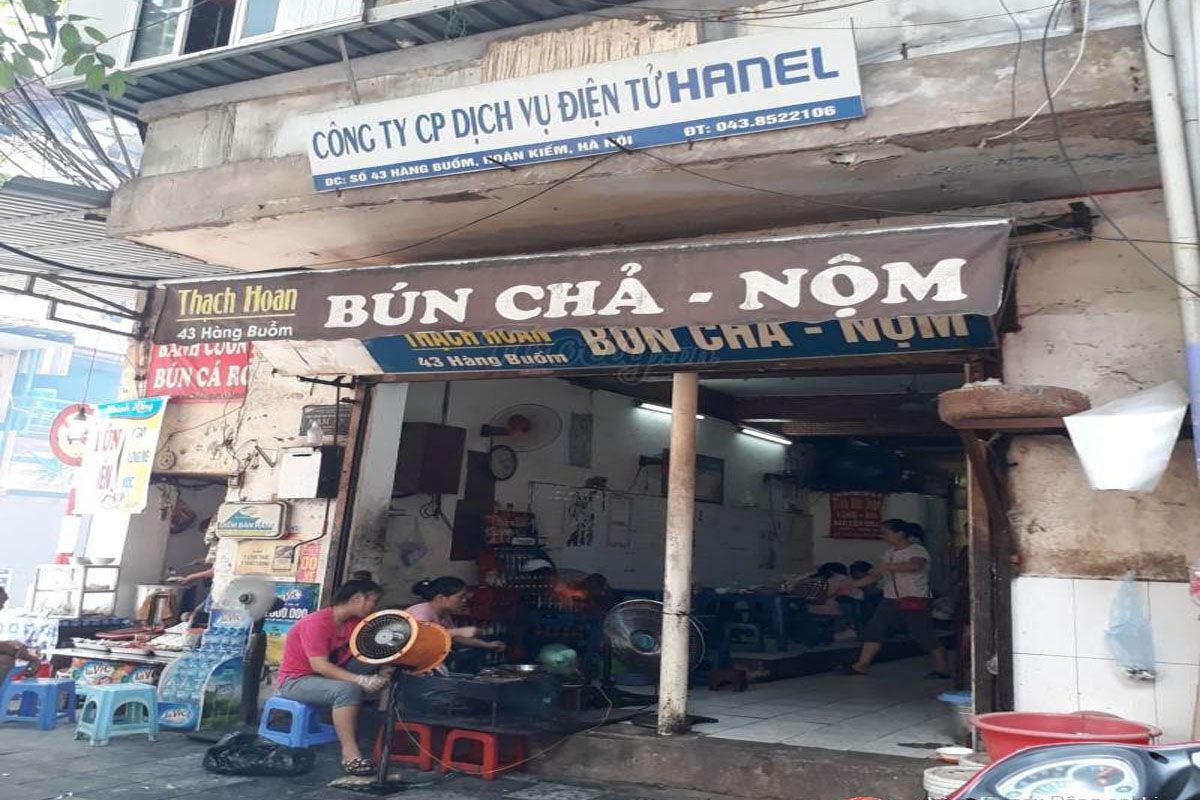}
  \caption{Example of North side signboard in Vietnam.}
  \label{North}
\end{figure}

Signboards from northern Vietnam often reflect a more formal and conservative communication style, rooted in traditional cultural norms and the historical role of the North as Vietnam’s political and cultural center. These signboards tend to be minimalistic in design, using straightforward, factual language with fewer visual distractions. They often avoid decorative elements and focus instead on clarity and direct messaging. For example, businesses may use names like "Bia Hà Nội"(Hanoi beer) or "Bún chả-nộm"(Bun cha-salad) see in ~\ref{North}, which clearly state the product or service being offered without embellishment.

This design style is influenced by the more reserved lifestyle and slower pace of urban development in certain northern areas, where brand identity is often built on trust, tradition, and clarity rather than visual spectacle. As a result, the language used on these signboards tends to be standard Vietnamese, with minimal use of slang or foreign words, making it relatively easier for models to process linguistically—but sometimes harder in cases where brevity reduces context.
\subsection{Southern Signboards}
In contrast, signboards in the southern regions—particularly in major cities like Ho Chi Minh City—tend to be more vibrant, commercial, and visually dense. These signboards often blend Vietnamese and English, employ multiple font styles, and contain rich visual elements that reflect a more dynamic and market-oriented lifestyle. This difference stems from the rapid economic growth, high population density, and diverse business landscape in the South, which drive a need for competitive visual branding to attract customers in busy commercial areas.

\begin{figure}[htp]
  \centering
\includegraphics[width=0.49\textwidth]{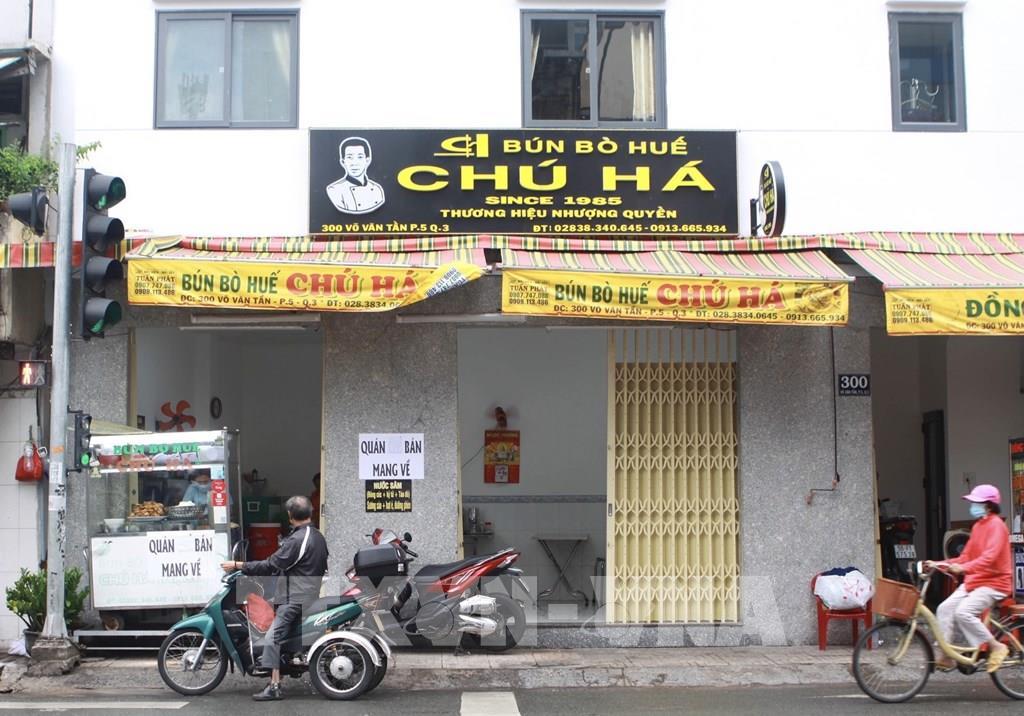}
\includegraphics[width=0.49\textwidth]{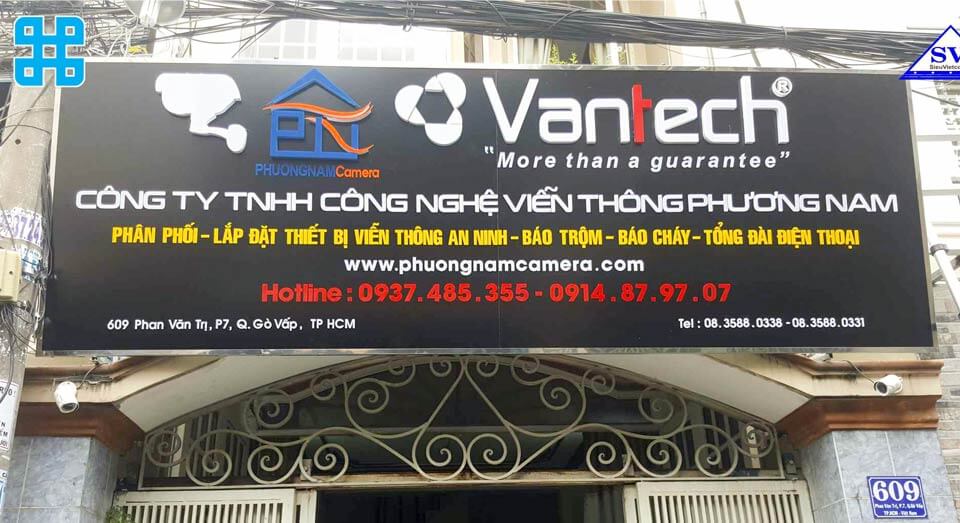}
  \caption{Example of South side signboard in Vietnam.}
  \label{South}
\end{figure}

Southern signboards often contain multiple layers of information, including names, slogans, contact info, promotions, and even hashtags or QR codes—all within a single frame.This advertising-heavy approach, while engaging to human viewers, can confuse VQA models due to textual clutter, non-standard language, and foreign word inclusion, especially if the model has not been exposed to such patterns during training.Example show in ~\ref{South}
\subsection{Result on Region Area Signboards}
\begin{figure}[htp]
  \centering
\includegraphics[width=0.49\textwidth]{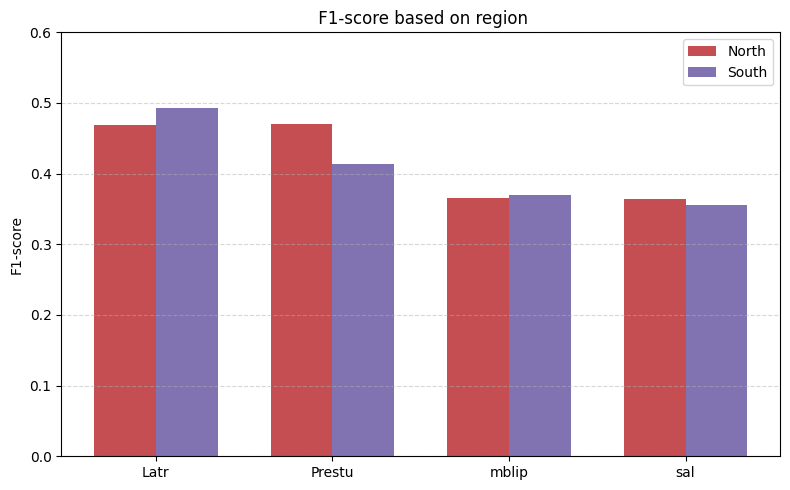}
\includegraphics[width=0.49\textwidth]{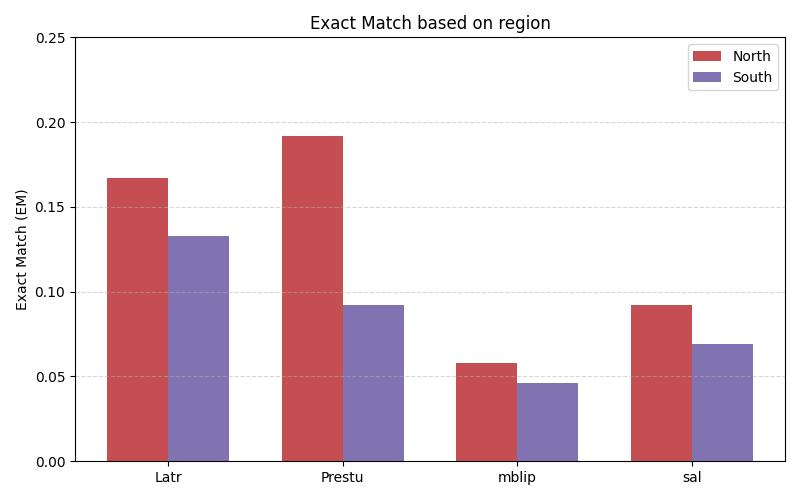}
  \caption{Result based on region signboard in Vietnam.}
  \label{Res}
\end{figure}

Figure ~\ref{Res} illustrates the F1-scores for four baseline models (Latr, Prestu, mblip, and sal) across the North and South. Notably, Latr achieved better performance in the South (0.493 vs. 0.469), suggesting its adaptability to the more dynamic and commercialized language style often found in southern signboards. In contrast, Prestu performed better in the North (0.470 vs. 0.414), where the language is typically more formal, concise, and easier to interpret. mblip and sal demonstrated minor differences between regions, indicating either general robustness or a lack of sensitivity to regional linguistic features.

Beside that with Em scores. All models showed higher EM scores in the North. This outcome is consistent with the observation that northern signboards often contain standard Vietnamese with shorter, more direct phrasing, making it easier for models to reproduce exact target answers. In contrast, southern signboards frequently incorporate English loanwords, slogans, and dense textual layouts, which introduce greater variability in possible answers and reduce the likelihood of exact matches.

These results highlight a critical challenge in Vietnamese VQA: regional linguistic and visual diversity directly impacts model generalization. Models like Prestu, which rely heavily on textual clarity, benefit from the formal style of northern signage. Conversely, models such as Latr may better exploit rich visual-textual cues common in southern advertising culture. The consistent EM gap across all models further suggests that current VQA systems are less equipped to handle linguistic variation and visual clutter typical of the southern region.

This regional disparity underscores the need for more balanced training data and model designs that account for sociolinguistic variation, especially in low-resource languages with strong regional dialects like Vietnamese.


\section{Conclusion and Future Work}
\label{sec9}
In this work, we present ViSignVQA, a novel dataset tailored for Visual Question Answering (VQA) on Vietnamese signboards. The dataset contains 10,762 signboard images and 25,573 question–answer pairs, capturing the unique characteristics of text-rich visual scenes in Vietnam. By introducing this resource, we aim to broaden the diversity of multilingual VQA benchmarks and provide a valuable foundation for advancing research in both the global VQA community and Vietnamese language processing.

Furthermore, we benchmarked a set of leading VQA approaches using ViSignVQA. Although these models exhibit competitive performance on widely adopted English VQA datasets, their accuracy dropped significantly on our benchmark, highlighting the linguistic and domain-specific challenges posed by Vietnamese signboards. This motivated us to conduct a comprehensive analysis of their performance, highlighting the limitations that emerged during experimentation and providing insights that can guide future research. Our analysis further revealed that incorporating OCR text significantly improves performance, leading to notable gains in the accuracy of baseline models for Vietnamese.
Besides that, we also evaluated the dataset using large language models (LLMs) via prompt-based fine-tuning approaches. In addition, we enhanced the multi-agent VQA framework by incorporating an OCR-guided module, allowing agents to better utilize textual content from signboards during the answer generation process.

Looking ahead, we recognize that the ViSignVQA dataset has broad potential beyond its application to the VQA task. One promising direction is its use in Visual Question Generation (VQG), where the objective is to automatically generate meaningful questions about an image given a specific answer \cite{zhang2017automatic, fan2018reinforcement, xu2020radial}. We plan to explore this task using our dataset, further extending its utility for the research community.

To enhance the applicability for signboard-related tasks, we plan to develop a chatbot designed to answer image-based questions, drawing inspiration from models like Flamingo \cite{alayrac2022flamingo}, GPT-4 \cite{openai2023gpt4}, and Gemini \cite{team2023gemini}. Built upon ViSignVQA as the main training resource and enriched with additional Vietnamese multimodal datasets, the chatbot is designed to robustly process the rich textual and visual information characteristic of signboards. We aim to create a practical and user-friendly application that allows for natural and flexible interaction in Vietnamese, tailored specifically for interpreting and responding to signboard-related queries.

\section*{Acknowledgements}
This research was supported by The VNUHCM-University of Information Technology’s Scientific Research Support Fund.

\section*{Declarations}

\textbf{Conflict of interest} The authors declare that they have no conflict of interest.

\section*{Data Availability}

Data will be made available on reasonable request.

\section*{Author Contribution}
Hieu Minh Nguyen: Conceptualization; Data curation; Formal analysis; Investigation; Methodology; Validation; Visualization; Writing - original draft.

Tam Le-Thanh Dang: Conceptualization; Data curation; Formal analysis; Investigation; Validation; Visualization; Writing - original draft.

Kiet Van Nguyen: Conceptualization; Formal analysis; Investigation; Methodology; Validation; Supervision; Writing - review\&editing.




\clearpage

\renewcommand\refname{References}
\bibliography{sn-bibliography}
\end{document}